\documentclass[journal]{IEEEtranTIE}
\usepackage{graphicx}
\usepackage{cite}
\usepackage{hyperref}
\usepackage{amsmath}
\usepackage{mathtools}
\usepackage{amssymb}
\usepackage{amsthm}
\usepackage{epsfig}
\usepackage{booktabs}
\usepackage{multirow}
\usepackage{makecell}
\usepackage{algpseudocode}
\usepackage{algorithm}
\usepackage{subfloat}
\usepackage{graphicx}
\usepackage{subfig}
\usepackage{color}
\newtheorem{definition}{\bf Definition}

\begin{document}
\title{Hyperbolic Binary Neural Network}
\author{Jun~Chen,
	Jingyang~Xiang,
	Tianxin~Huang,
	Xiangrui~Zhao,
	and Yong~Liu,~\IEEEmembership{Member,~IEEE}
    \thanks{This work was supported by the National Natural Science Foundation of China under Grant 62103363. \textit{Corresponding authors: Jun~Chen and Yong~Liu} (e-mail: junc@zju.edu.cn; yongliu@iipc.zju.edu.cn).}
    \thanks{Jun~Chen is with the National Special Education Resource Center for Children with Autism, Zhejiang Normal University, Hangzhou 311231, China, and with the Institute of Cyber-Systems and Control, Zhejiang University, Hangzhou 310027, China, and also with the School of Computer Science and Technology, Zhejiang Normal University, Jinhua 321004, China.}
    \thanks{Jingyang~Xiang, Tianxin~Huang, Xiangrui~Zhao and Yong~Liu are with the Institute of Cyber-Systems and Control, Zhejiang University, Hangzhou 310027, China.}
}

\maketitle
	
\begin{abstract}
Binary Neural Network (BNN) converts full-precision weights and activations into their extreme 1-bit counterparts, making it particularly suitable for deployment on lightweight mobile devices.
While binary neural networks are typically formulated as a constrained optimization problem and optimized in the binarized space, general neural networks are formulated as an unconstrained optimization problem and optimized in the continuous space.
This paper introduces the Hyperbolic Binary Neural Network (HBNN) by leveraging the framework of hyperbolic geometry to optimize the constrained problem.
Specifically, we transform the constrained problem in hyperbolic space into an unconstrained one in Euclidean space using the Riemannian exponential map.
On the other hand, we also propose the Exponential Parametrization Cluster (EPC) method, which, compared to the Riemannian exponential map, shrinks the segment domain based on a diffeomorphism. This approach increases the probability of weight flips, thereby maximizing the information gain in BNNs.
Experimental results on CIFAR10, CIFAR100, and ImageNet classification datasets with VGGsmall, ResNet18, and ResNet34 models illustrate the superior performance of our HBNN over state-of-the-art methods.
\end{abstract}


\markboth{IEEE TRANSACTIONS ON NEURAL NETWORKS AND LEARNING SYSTEMS}%
{}

\section{Introduction}
\label{sec1}

\IEEEPARstart{D}{eep} Neural Networks (DNNs) have achieved remarkable success in various computer vision fields, including image classification~\cite{krizhevsky2012imagenet,he2016deep}, object detection~\cite{redmon2016you,he2017mask}, semantic segmentation~\cite{long2015fully,noh2015learning}, and more.
However, the massive parameters and computational complexity of DNNs, which contribute to their success, limit their deployment on lightweight mobile devices.
To address this problem, various compression methods are being proposed, with the main approaches including pruning~\cite{ding2019global,lin2020hrank,bai2023unified}, quantization~\cite{banner2018scalable,helwegen2019latent,chen2020propagating,chen2023learning,chen2023data}, and distillation~\cite{liu2023dccd}.

In the context of resource-constrained and low-power devices, quantization emerges as a more effective and universal scheme compared to pruning~\cite{9056829}.
Specifically, quantization converts full-precision weights and activations into their low-precision counterparts.
In the extreme case, neural network binarization restricts weights and activations to two possible discrete values $\{-1,+1\}$, offering two advantages: (1) a 32$\times$ reduction in memory compared to the corresponding full-precision version; (2) the multiply-accumulation operation can be replaced with the efficient xnor and bitcount operations.

\begin{figure}
	\centering
	\includegraphics[trim=50 40 20 40, scale=0.35]{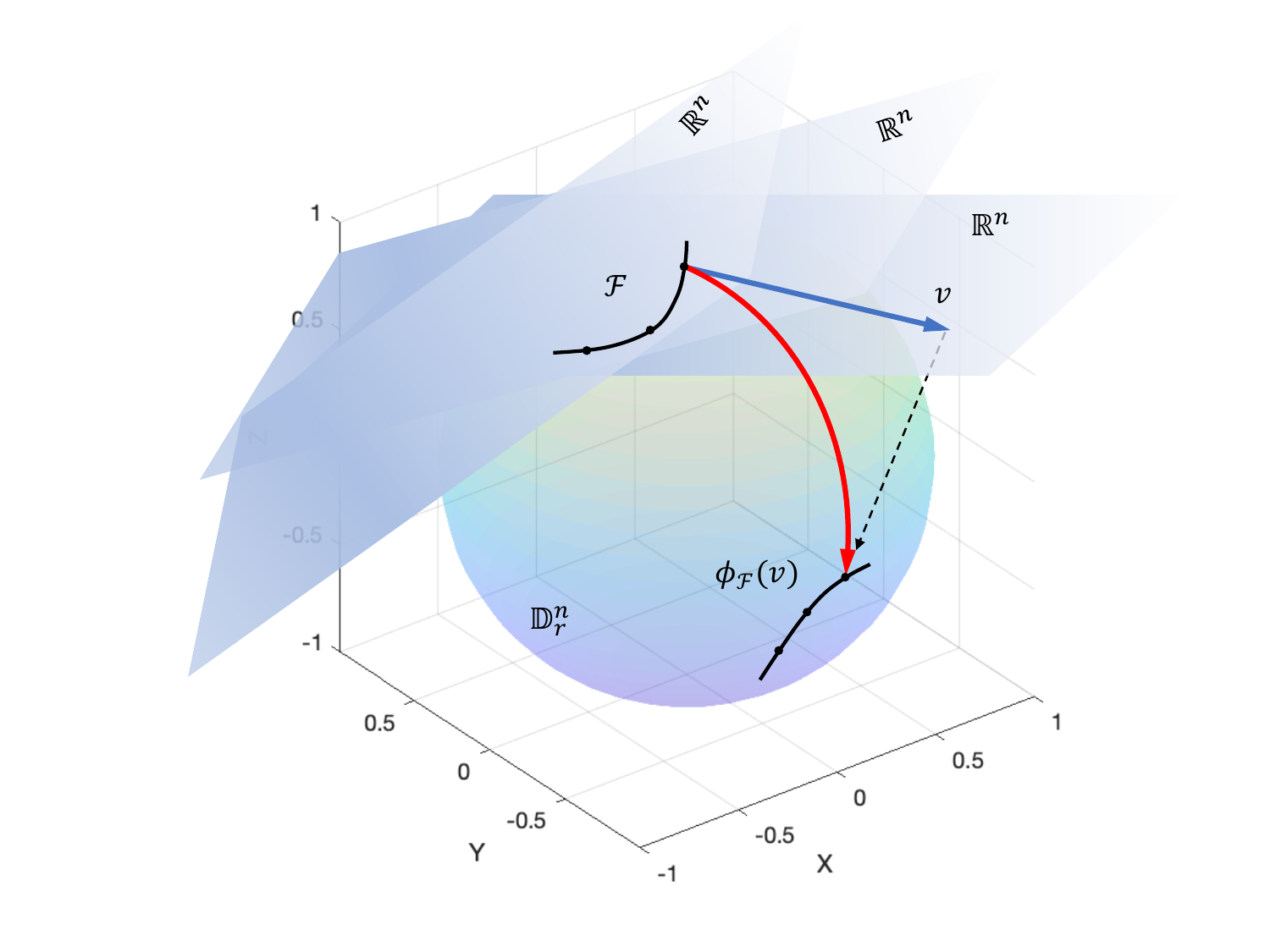}
	\caption{The exponential parametrization cluster $\phi_{\mathcal{F}}$ transforms a vector $v$ into the mapped cluster $\phi_{\mathcal{F}}(v)$ using an original cluster $\mathcal{F}=\{\mathcal{F}_1,\mathcal{F}_2,\cdots,\mathcal{F}_t\}$, where $\mathcal{F}$ and $\phi_{\mathcal{F}}(v)$ exist in hyperbolic space, while $v$ resides in Euclidean space. In contrast, the Riemannian exponential map $\exp$ transforms a vector $v$ into the mapped point $\exp(v)$.}
	\label{fig1}
\end{figure}

Neural network binarization is typically formulated as a constrained optimization problem with respect to the dataset $\mathcal{D}=\{\mathbf{x}_{i}, \mathbf{y}_{i}\}^m_{i=1}$ and the set of all possible binarized solutions $\mathcal{X} \subset \mathbb{R}^n$:
\[
	\min _{\mathbf{w} \in \mathcal{X}} \mathcal{L}(\mathbf{w} ; \mathcal{D}):=\frac{1}{m} \sum_{i=1}^{m} \mathcal{L}\left(\mathbf{w} ;\left(\mathbf{x}_{i}, \mathbf{y}_{i}\right)\right),
\]
where $\mathbf{w}$ is an $n$ dimensional weight vector, and $\mathcal{L}$ represents the loss function, such as cross-entropy loss.
Using a mirror descent framework~\cite{bubeck2015convex}, Ajanthan \emph{et al.}~\cite{ajanthan2021mirror} transformed the constrained problem into an unconstrained one through a mapping $P: \mathbb{R}^n \rightarrow \mathcal{X}$ such that
\[
	\min _{\tilde{\mathbf{w}} \in \mathbb{R}^{n}} \mathcal{L}(P(\tilde{\mathbf{w}}) ; \mathcal{D}).
\]
Subsequently, $P(\tilde{\mathbf{w}}) \in \mathcal{X}$ is gradually binarized to a discrete set $\mathcal{B}^n=\{-1,+1\}^n$ during the training process, where $P$ is defined as a mirror map.

In BNNs, the norm of the binarized weight vector in each layer is fixed and is solely determined by the dimension of the weight.
In other words, the binarized weight resides on a ball with a constant radius, forming a hyperbolic space.
In this paper, we introduce a Hyperbolic Binary Neural Network (HBNN) to formulate neural network binarization as a optimization problem in the framework of hyperbolic space.
Specifically, we transform the constrained problem in hyperbolic space into an unconstrained one in Euclidean space using the Riemannian exponential map.
This approach is more conducive to optimizing BNNs than directly converting the optimization problem from the constrained and binarized space to the unconstrained and continuous space.

On the other hand, recent research~\cite{lin2020rotated} has demonstrated that a high ratio of weight flips, where weight flips mean that positive values turn to negative values and vice versa, can maximize the information gain to optimize BNNs' performance. In this context, we propose the Exponential Parametrization Cluster ($\phi_{\mathcal{F}}(\cdot): \mathbb{R}^n \rightarrow \mathbb{D}^n_r$) shown in Figure~\ref{fig1}. This approach is a differentiable map from the tangent space ($\mathbb{R}^n$) to the hyperbolic space ($\mathbb{D}^n_r$). In this case, the constrained optimization problem in hyperbolic space is transformed into an unconstrained one in Euclidean space:
\begin{equation}
	\begin{aligned}
		&\text{Original problem:} \min _{\mathbf{w} \in \mathbb{D}^n_r} \mathcal{L}(\mathbf{w} ; \mathcal{D}), \\ & \text{Unconstrained problem:} \min _{\tilde{\mathbf{w}} \in \mathbb{R}^{n}, \mathcal{F} \in \mathbb{D}^n_r} \mathcal{L}(\phi_{\mathcal{F}}(\tilde{\mathbf{w}}) ; \mathcal{D}),
	\end{aligned}
	\label{problem}
\end{equation}
where the cluster $\mathcal{F}$ consists of a series of candidate points $\{\mathcal{F}_1,\mathcal{F}_2,\cdots,\mathcal{F}_t\}$. In comparison to the Riemannian exponential map~\cite{guggenheimer2012differential} $\exp(\cdot)$, our proposed Exponential Parametrization Cluster (EPC) extends the mapping result from a single point to a cluster of points. Inherently, the Riemannian exponential map $\exp(\cdot)$ is equivalent to $\phi_{\mathcal{F}_i}(\cdot)$, where $\mathcal{F}_i$ is a candidate point from the cluster $\mathcal{F}$.

The main contributions of this paper are summarized in the following three aspects:
\begin{enumerate}
    \item We propose the Hyperbolic Binary Neural Network by leveraging the framework of hyperbolic geometry to optimize the constrained problem. Specifically, we transform the constrained problem in hyperbolic space into an unconstrained one in Euclidean space using the Riemannian exponential map.
    \item We introduce the exponential parametrization cluster, which, compared to the Riemannian exponential map, shrinks the segment domain on the basis of a diffeomorphism. This approach increases the probability of weight flips, maximizing the information gain in BNNs.
    \item Experimental results on CIFAR10, CIFAR100, and ImageNet classification datasets with VGGsmall, ResNet18, and ResNet34 models illustrate the superior performance of our HBNN over state-of-the-art methods.
\end{enumerate}

\section{Related Work}

\textbf{Optimization on Manifolds.} Many optimization methods on manifolds have Riemannian analogs~\cite{absil2009optimization,chen2024decentralized}. Parametrization is an important technique for converting problems with manifold constraints into unconstrained problems in Euclidean space.
Helfrich \emph{et al.}~\cite{helfrich2018orthogonal} introduced orthogonal and unitary Cayley parametrizations, which construct orthogonal weight matrices through a scaled Cayley transform in recurrent neural networks.
Lezcano-Casado \emph{et al.}~\cite{lezcano2019cheap} introduced the orthogonal exponential parametrization derived from Lie group theory using the Riemannian exponential map.
Lezcano-Casado~\cite{lezcano2019trivializations} further introduced dynamic parametrization as a gradient-based optimization that combines the advantages of the Riemannian exponential and Lie exponential.

\textbf{Binarization Methods.} The introduction of the non-differentiable sign function in neural network binarization leads to a performance drop. 
For instance, XNOR~\cite{rastegari2016xnor} introduced accurate approximations by binarizing not only the weights but also the intermediate representations in DNNs. This approach aims to reduce the quantization error between the full-precision weights and their binarized counterparts.
XNOR++~\cite{bulat2019xnor} further fused the activation and weight scaling factors into a single factor, improving overall performance.
BiReal~\cite{liu2020bi} addressed the problem of infinite or zero gradients caused by the sign function by propagating full-precision activations through a parameter-free shortcut in each binarized convolution.
Proxy-BNN~\cite{he2020proxybnn} introduced a proxy matrix to serve as the basis for the latent parameter space, aiming to reduce the quantization error of weights and restore the smoothness of BNNs.
Recently, IR-Net~\cite{qin2020forward} proposed a balanced and standardized binarization method in the forward pass, minimizing the information loss by maximizing the information entropy of binarized weights and minimizing the quantization error.
RBNN~\cite{lin2020rotated} analyzed the angle alignment between full-precision weights and their binarized counterparts, highlighting that around 50\% weight flips can maximize the information gain.
ReCU~\cite{xu2021recu} employed the weight normalization~\cite{salimans2016weight,huang2017centered} to revive ``dead weights", increasing the probability of updating these weights in BNNs.

\section{Preliminaries}

Here, we provide background knowledge on Riemannian geometry and BNNs.

\subsection{Riemannian Geometry}

We briefly introduce the basic concepts of Riemannian geometry, and for more in-depth propositions, see~\cite{petersen2006riemannian,guggenheimer2012differential}.

\textbf{Tangent Space.} For an $n$-dimensional connected manifold $\mathcal{M}$, the tangent space at a point $p \in \mathcal{M}$ is defined as $T_p \mathcal{M}$. This is a real vector space that can be described as a high-dimensional generalization of a tangent plane. And such a tangent space exists for all points $p \in \mathcal{M}$.
Thus, the description of tangent spaces aligns with Euclidean space, denoted as $T_p \mathcal{M} \cong \mathbb{R}^n$.

\textbf{Riemannian Manifold.} Riemannian manifolds are endowed with a smooth metric $g_p: T_p \mathcal{M} \times T_p \mathcal{M} \rightarrow \mathbb{R}$ that varies smoothly with $p$, enabling the construction of a distance function $d_g: \mathcal{M} \times \mathcal{M} \rightarrow \mathbb{R}$. When describing a Riemannian manifold, the Riemannian metric is inherently equipped by default, denoted as $(\mathcal{M}, g)$.

\textbf{Geodesics.} In a complete Riemannian manifold, a smooth path of minimal length between two points on $\mathcal{M}$ is termed a geodesic. Mathematically, a geodesic is defined as $\gamma_{p,v}(t): t \in [0,1] \rightarrow \mathcal{M}$ such that $\gamma_{p,v}(0)=p$, $\gamma'_{p,v}(0)=v$ for $v \in T_p \mathcal{M}$. Geodesics serve as the generalization of straight lines in Euclidean space.

\textbf{Exponential Map.} The Riemannian exponential map, denoted as $\exp: T_p \mathcal{M} \rightarrow \mathcal{M}$, serves to map rays starting at the origin in the tangent space $T_p \mathcal{M}$ to geodesics on $\mathcal{M}$. For a given geodesic, the parameter $t$ ranges from $0$ to $1$, resulting in $\exp(tv):=\gamma_{p,v}(t)$. Specifically, the distance on the manifold between a point $p$ and the exponential map $\exp(v)$ is given by $d_g(p, \exp(v))=\Vert v \Vert_g$.

\subsection{Binary Neural Network}

\begin{figure*}[t]
	\centering
	\includegraphics[trim=30 60 30 40,scale=0.6]{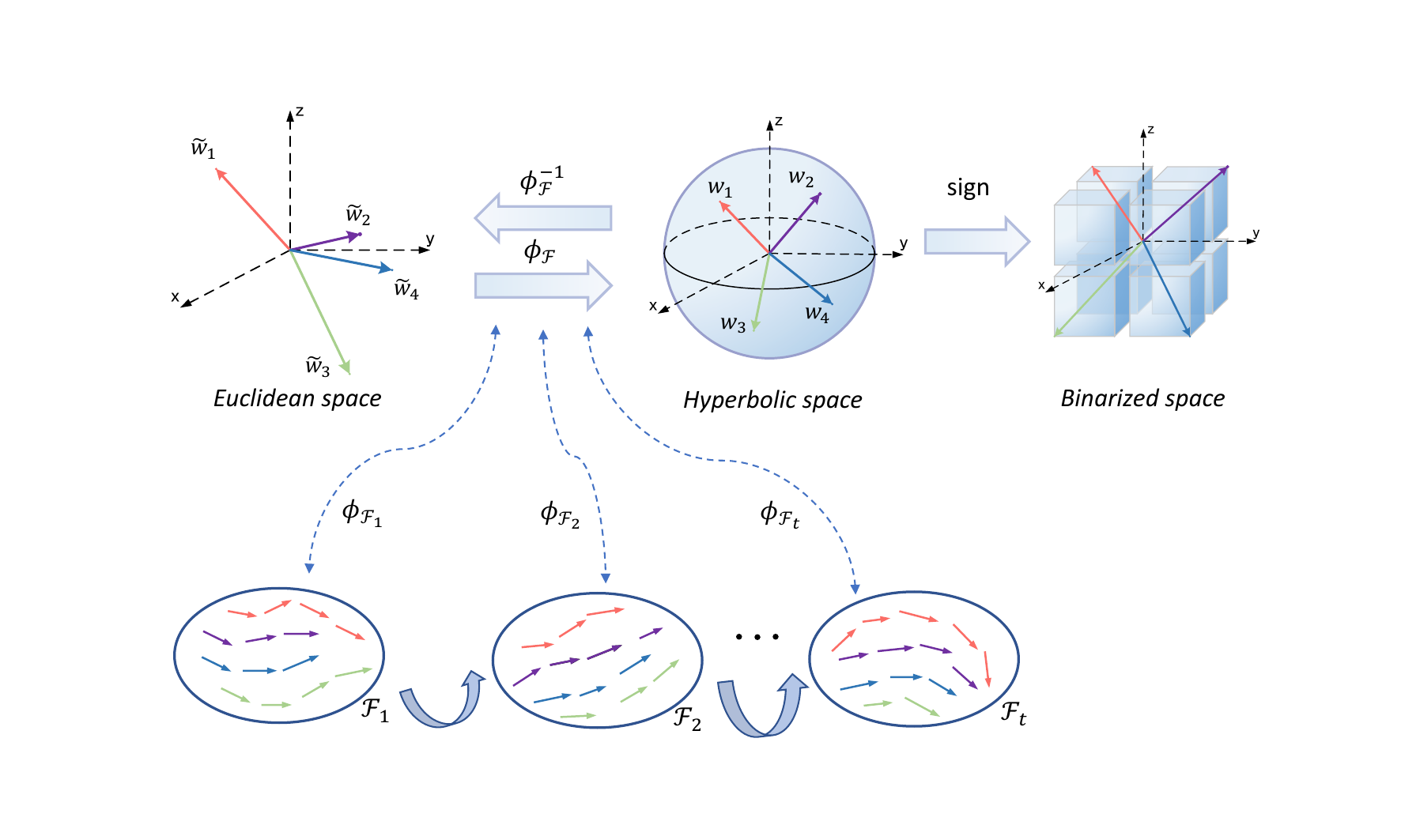}
	\caption{The overview of our HBNN with the EPC. By training an original cluster $\mathcal{F}=\{\mathcal{F}_1,\mathcal{F}_2,\cdots,\mathcal{F}_t\}$, we map a weight vector $\tilde{\mathbf{w}}$ into the mapped cluster $\phi_{\mathcal{F}}(\tilde{\mathbf{w}})=\{\phi_{\mathcal{F}_1}(\tilde{\mathbf{w}}),\phi_{\mathcal{F}_2}(\tilde{\mathbf{w}}),\cdots,\phi_{\mathcal{F}_t}(\tilde{\mathbf{w}})\}$. Subsequently, we obtain an optimal exponential parametrization (Let's assume $\phi_{\mathcal{F}_i}(\cdot)$) based on the mapped cluster. Consequently, we continue to optimize the weight vector $\tilde{\mathbf{w}}$ via $\phi_{\mathcal{F}_i}(\cdot)$. Note that HBNN obtains the binarized weight vector via  $\operatorname{sign}(\phi_{\mathcal{F}_i}(\tilde{\mathbf{w}}))$.}
	\label{main}
\end{figure*}

Now, let's delve into the mechanism of BNNs and explore how the binarization and gradients are computed.

\textbf{Forward Pass.} During the inference phase of a BNN, the binarization function is expressed in a deterministic form~\cite{courbariaux2016binarized,rastegari2016xnor}:
\begin{equation}
	x^{b}=\operatorname{sign}(x)= \begin{cases}+1 & \text { if } x \geq 0 ,\\ -1 & \text {otherwise,}\end{cases}
	\label{sign}
\end{equation}
where $x$ can represent either weights $\mathbf{w}$ or activations $\mathbf{a}$.

\textbf{Backward Pass.} During back-propagation, the gradient suffers from the problem of either infinite or zero when propagating through the binarization function. To address this problem, Hinton and Bengio~\cite{hinton2012neural,bengio2013estimating} proposed the Straight-Through Estimator. Consequently, this estimator of the gradient with respect to binarized weights can be approximated as
\begin{equation}
	\frac{\partial \mathcal{L}}{\partial \mathbf{w}}=\frac{\partial \mathcal{L}}{\partial \mathbf{w}^b}\cdot\frac{\partial \mathbf{w}^b}{\partial \mathbf{w}},\;\; \frac{\partial \mathbf{w}^b}{\partial \mathbf{w}}:=\left\{\begin{array}{ll}
		1 & \text { if } \quad|\mathbf{w}| \leq 1 \\
		0 & \text { otherwise}
	\end{array}\right. .
	\label{ste}
\end{equation}
On the other hand, base on the polynomial function~\cite{liu2020bi}, an estimator of the gradient with respect to binarized activations can be formulated as
\begin{equation}
	\frac{\partial \mathcal{L}}{\partial \mathbf{a}}=\frac{\partial \mathcal{L}}{\partial \mathbf{a}^b}\cdot\frac{\partial \mathbf{a}^b}{\partial \mathbf{a}},\;\; \frac{\partial \mathbf{a}^b}{\partial \mathbf{a}}:=\begin{cases}2+2 \mathbf{a}, & \text { if }-1 \leq \mathbf{a}<0 \\ 2-2 \mathbf{a}, & \text { if } 0 \leq \mathbf{a}\leq 1 \\ 0, & \text { otherwise }\end{cases} .
	\label{ste2}
\end{equation}

\textbf{Activation Function.} In a BNN, the activation function such as \emph{ReLU} are avoided because the binarized activation values through \emph{ReLU} would all become $1$. Typically, \emph{Hardtanh} is applied instead.

\section{Hyperbolic Binary Neural Network}
\label{sec3}

\subsection{The Poincar\'e Ball}

The hyperbolic space has several isometric models~\cite{anderson2006hyperbolic}, which are not only conformal to Euclidean space but also offer powerful and meaningful geometrical representations~\cite{ganea2018hyperbolic}. We choose the Poincar\'e ball model, as suggested by the previous works~\cite{nickel2017poincare,ganea2018hyperbolic1}.
By denoting an $n$-dimensional Poincar\'e ball with radius $1/\sqrt{r}$ as $\mathbb{D}_{r}^{n}:=\left\{x \in \mathbb{R}^{n} \mid r\|x\|^{2}<1\right\}$, the equipped hyperbolic metric is given by:
\begin{equation}
	g_{x}^{H}=\lambda_{x}^{2} g^{E}, \quad \text { where } \lambda_{x}:=\frac{2}{1-r\|x\|^{2}}.
\end{equation}
Here, $g^E$ represents the Euclidean metric, i.e., the identity matrix.
For $r>0$, $\mathbb{D}_{r}^{n}$ denotes the open ball (Poincar\'e ball).
When the radius $r$ equals to zero, the Poincar\'e ball $\mathbb{D}_{r}^{n}$ recovers the Euclidean space, i.e., $\mathbb{D}_{0}^{n}=\mathbb{R}^n$.
Similarly, we can denote an $n$-dimensional sphere with radius $1/\sqrt{r}$ as $\mathbb{S}_{r}^{n}:=\left\{x \in \mathbb{R}^{n} \mid r\|x\|^{2}=1\right\}$, expressed by the boundary of the Poincar\'e ball, namely $\partial\mathbb{D}_{r}^{n}$.

\subsection{Exponential Parametrization Cluster (EPC)}

Building upon Eq.(\ref{problem}), we aim to transform the constrained optimization problem of binarization in hyperbolic space into an unconstrained optimization problem in Euclidean space.
For a weight vector $\tilde{\mathbf{w}}$ in Euclidean space, we can compute its EPC, i.e., $\phi_{\mathcal{F}}(\tilde{\mathbf{w}})$, which is composed of a series of weight vectors $\{\phi_{\mathcal{F}_1}(\tilde{\mathbf{w}}),\phi_{\mathcal{F}_2}(\tilde{\mathbf{w}}),\cdots,\phi_{\mathcal{F}_t}(\tilde{\mathbf{w}})\}$ in hyperbolic space.

Given a weight vector $\tilde{\mathbf{w}} \in T_p \mathbb{D}_{r}^{n} (\cong \mathbb{R}^n) \backslash \{\textbf{0}\}$, where $p \in \mathbb{D}_{r}^{n}$, the EPC with a cluster $\mathcal{F}=\{\mathcal{F}_1,\mathcal{F}_2,\cdots,\mathcal{F}_t\} \in \mathbb{D}_{r}^{n}$ can be expressed in the Poincar\'e ball with the radius $1/\sqrt{r}$ as follows:
\begin{equation}
	\phi_{\mathcal{F}}(\cdot):= \left\{\begin{array}{c}
		\mathcal{F}_1 \oplus\left(\tanh \left(\sqrt{r} \frac{\lambda_{p}\|\cdot\|}{2}\right) \frac{\cdot}{\sqrt{r}\|\cdot\|}\right)  \\
		\mathcal{F}_2 \oplus\left(\tanh \left(\sqrt{r} \frac{\lambda_{p}\|\cdot\|}{2}\right) \frac{\cdot}{\sqrt{r}\|\cdot\|}\right) \\
		\vdots \\
		\mathcal{F}_t \oplus\left(\tanh \left(\sqrt{r} \frac{\lambda_{p}\|\cdot\|}{2}\right) \frac{\cdot}{\sqrt{r}\|\cdot\|}\right) 
	\end{array}\right\}.
	\label{exp}
\end{equation}
Geometrically, the EPC starts with a cluster $\mathcal{F}$ and takes $v$ as the initial tangent vector on the geodesic. This vector satisfies that the geodesic distance from the mapped cluster $\phi_{\mathcal{F}}(v)$ to the original cluster is $\Vert v \Vert_g$. 
It's important to note that the notation $\oplus$ used here follows the addition formalism for hyperbolic geometry, differing from the traditional Euclidean geometry.
The non-associative algebra for hyperbolic geometry can be expressed in the framework of gyrovector spaces~\cite{ungar2001hyperbolic,ungar2008gyrovector}.

\textbf{Addition~\cite{ganea2018hyperbolic}.} In the Poincar\'e ball, the addition of $p$ and $q$ in $\mathbb{D}_{r}^{n}$ is defined as:
\begin{equation}
	p \oplus q:=\frac{\left(1+2 r\langle p, q\rangle+r\|q\|^{2}\right) p+\left(1-r\|p\|^{2}\right) q}{1+2 r\langle p, q\rangle+r^{2}\|p\|^{2}\|q\|^{2}}.
\end{equation}

Given that the Riemannian exponential map $\exp(\cdot)$ is constrained by a point, the corresponding representations $\exp(\tilde{\mathbf{w}})$ do not contribute to an increased probability of weight flips.
In contrast, our mapped cluster $\phi_{\mathcal{F}}(\tilde{\mathbf{w}})$ provides more candidate representations by training a cluster $\mathcal{F}$, thereby increasing the probability of weight flips. We will theoretically elaborate on the role of the EPC in weight flips in Section~\ref{sec5}. An overview of our HBNN with the EPC is presented in Figure~\ref{main}.

Subsequently, we can formulate the unconstrained problem for the weight vector, unifying Eq.(\ref{problem}) and Eq.(\ref{exp}), as follows:
\begin{equation}
	\min_{\tilde{\mathbf{w}} \in \mathbb{R}^{n}} \min_{\mathcal{F} \in \mathbb{D}_{r}^{n}}\mathcal{L}\left(\{\phi_{\mathcal{F}_1}(\tilde{\mathbf{w}}),\phi_{\mathcal{F}_2}(\tilde{\mathbf{w}}),\cdots,\phi_{\mathcal{F}_t}(\tilde{\mathbf{w}})\}; \mathcal{D}\right).
	\label{hbnn}
\end{equation}

\subsection{Backward Mode and Gradient Computation}

In order to fully implement our HBNN in the deep learning framework, it is crucial to efficiently compute gradients for the problem stated in Eq.(\ref{hbnn}).
During back-propagation, we first keep the weight vector $\tilde{\mathbf{w}}$ unchanged. Using a learning rate $\eta>0$, we then update the cluster $\mathcal{F}$ in hyperbolic space:
\begin{equation}
	\mathcal{F} \leftarrow \left\{\begin{array}{c}
		\mathcal{F}_1 \oplus-\eta \otimes \frac{\partial \mathcal{L}}{\partial \mathcal{F}_1}  \\
		\mathcal{F}_2 \oplus-\eta \otimes \frac{\partial \mathcal{L}}{\partial \mathcal{F}_2} \\
		\vdots \\
		\mathcal{F}_t \oplus-\eta \otimes \frac{\partial \mathcal{L}}{\partial \mathcal{F}_t}
	\end{array}\right\},
	\label{p}
\end{equation}
where the notation $\otimes$ represents the multiplication formalism for hyperbolic geometry.

\textbf{Multiplication~\cite{ganea2018hyperbolic}.} In the Poincar\'e ball, the scalar multiplication of $p \in \mathbb{D}_{r}^{n} \backslash \{\textbf{0}\}$ by $c \in \mathbb{R}$ is defined as:
\begin{equation}
	c \otimes p:=(1 / \sqrt{r}) \tanh \left(c \tanh ^{-1}(\sqrt{r}\|p\|)\right) \frac{p}{\|p\|}.
\end{equation}

Recall that the Straight-Through Estimator $\partial \mathcal{L} / \partial \mathbf{w}=\partial \mathcal{L} / \partial \operatorname{sign}(\mathbf{w})$ holds when $|\mathbf{w}| \leq 1$ is satisfied, as indicated by Eq.(\ref{ste}).
In hyperbolic space, the weight vector $\mathbf{w}:=\phi_{\mathcal{F}}(\tilde{\mathbf{w}}) \in \mathbb{D}_{r}^{n}$ naturally satisfies the constraint $\Vert\mathbf{w}\Vert < 1/\sqrt{r}$.
By slightly modifying the bounds of the Straight-Through Estimator ($1 \rightarrow 1/\sqrt{r}$), we can directly use $\partial \mathcal{L} / \partial \mathbf{w}=\partial \mathcal{L} / \partial \operatorname{sign}(\mathbf{w})$, which is always guaranteed to hold.

Assuming that $\mathcal{F}_i$ is an optimal point ($\phi_{\mathcal{F}_i}(\cdot)$ represents an optimal exponential parametrization) obtained by updating Eq.(\ref{p}), we have
\begin{equation}
	\min_{\tilde{\mathbf{w}} \in \mathbb{R}^{n}} \mathcal{L}\left(\phi_{\mathcal{F}_i}(\tilde{\mathbf{w}}); \mathcal{D}\right).
\end{equation}
Following the Straight-Through Estimator, we can compute the gradients $\partial \mathcal{L} / \partial \mathbf{w}$ to update the weight vector in the unconstrained Euclidean space:
\begin{equation}
	\tilde{\mathbf{w}} \leftarrow \tilde{\mathbf{w}} - \eta \frac{\partial \mathcal{L}}{\partial \mathbf{w}} \phi'_{\mathcal{F}_i}(\tilde{\mathbf{w}}).
	\label{w}
\end{equation}
Notably, the optimization of HBNN is an iterative process. Initially, we update the exponential parametrization cluster $\phi_{\mathcal{F}}(\cdot)$ to obtain the optimal exponential parametrization $\phi_{\mathcal{F}_i}(\cdot)$ while keeping the weight vector $\tilde{\mathbf{w}}$ fixed. Subsequently, we update the weight vector $\tilde{\mathbf{w}}$ using $\phi_{\mathcal{F}_i}(\cdot)$. The training process is summarized in Algorithm~\ref{alg}.

Intuitively, we can map the weight vector from hyperbolic space back to Euclidean space using the inverse of the optimal exponential parametrization $\phi_{\mathcal{F}_i}(\cdot)$. This mapping is a differentiable, as confirmed by the algebraic identity $\phi_{\mathcal{F}_i}^{-1}(\phi_{\mathcal{F}_i}(v))=v$, satisfying a closed-formula.

\begin{algorithm}[thbp]
	\caption{Forward and Backward Propagation of HBNN}
	\label{alg}
	\begin{algorithmic}[1]
		\Require
		A minibatch of data samples $\mathcal{D}=\{\mathbf{x}_{i}, \mathbf{y}_{i}\}^m_{i=1}$, current binary weight $\mathbf{w}^{b}_k$, latent full-precision unconstrained weight $\tilde{\mathbf{w}}_k$, latent full-precision constrained weight $\mathbf{w}_k$, the cluster $\mathcal{F}$, and a learning rate $\eta$.
		\Ensure
		Update $\tilde{\mathbf{w}}_k$ and $\mathcal{F}$.
		\State \{\textbf{Forward propagation}\}
		\For{$k=1$ to $l-1$}
		\State Compute the weight via an optimal exponential parametrization:  $\mathbf{w}_k \leftarrow \phi_{\mathcal{F}_i}(\tilde{\mathbf{w}}_k)$;
		\State Binarize the weight: $\mathbf{w}^{b}_k\leftarrow \operatorname{sign}(\mathbf{w}_k)$;
		\State Binarize the activation: $\mathbf{a}^{b}_{k-1}\leftarrow \operatorname{sign}(\mathbf{a}_{k-1})$;
		\State Perform: $\mathbf{a}_k\leftarrow\operatorname{XnorDotProduct}( \mathbf{w}^{b}_k,\mathbf{a}^{b}_{k-1})$;
		\State Perform: $\mathbf{a}_k\leftarrow\operatorname{BatchNorm}(\mathbf{a}_k)$;
		\EndFor
		\State \{\textbf{Backward propagation}\}
		\State Optimize the unconstrained problem with Eq.(\ref{hbnn});
		\State Compute the gradient of the overall loss function, i.e., $\frac{\partial \mathcal{L}}{\partial \mathbf{a}}$, $\frac{\partial \mathcal{L}}{\partial \mathbf{w}}$ and $\frac{\partial \mathcal{L}}{\partial \mathcal{F}}$, where the sign function can be handled in Eq.(\ref{ste}) for the weight and Eq.(\ref{ste2}) for activation;
		\State \{\textbf{The parameter update}\}
		\State Optimize the exponential parametrization cluster $\phi_{\mathcal{F}}(\cdot)$ in Eq.(\ref{exp}) by updating the cluster in Eq.(\ref{p}), then obtain the optimal exponential parametrization $\phi_{\mathcal{F}_i}(\cdot)$;
		\State Update the weight using Eq.(\ref{w}) based on the optimal exponential parametrization $\phi_{\mathcal{F}_i}(\cdot)$;
	\end{algorithmic}
\end{algorithm}

\section{Method Analysis}
\label{sec5}

\subsection{Theoretical Analysis}

In Riemannian geometry, the Riemannian exponential map serves as a metric change. In this paper, the exponential parametrization cluster $\phi_{\mathcal{F}}(\cdot)$ applies gradient descent to update the cluster $\mathcal{F}$, which is an operation equivalent to the Riemannian exponential map $\exp(\cdot)$ with an optimal metric change in hyperbolic space. This optimal metric change is evaluated by comparing multiple changes of metric, as opposed to a single change, in hyperbolic space.

\begin{definition}
	\label{thm1}
	\textbf{Diffeomorphism~\cite{anderson2006hyperbolic}.} Given a complete Riemannian manifold $(\mathcal{M},g)$ and a point $p \in \mathcal{M}$, the exponential map $\phi$ with respect to the largest convex open neighborhood of zero $\mathcal{X}_p \subseteq T_p \mathcal{M}$ is a diffeomorphism.
\end{definition}

According to Definition~\ref{thm1}, the exponential parametrization cluster is a diffeomorphism in the Poincar\'e ball $\mathbb{D}^n_r$. This property ensures that the optimization of parametrized weight vectors does not introduce or eliminate local minima at the loss landscape. However, the diffeomorphism ceases at the boundary of the Poincar\'e ball $\partial\mathbb{D}^n_r$, i.e., the sphere $\mathbb{S}^n_r$, potentially altering the local minima. Therefore, the Poincar\'e ball $\mathbb{D}^n_r$ is a preferable choice over the sphere $\mathbb{S}^n_r$.

\begin{definition}
	\label{thm2}
	\textbf{Segment~\cite{petersen2016riemannian}.} The segment domain $\operatorname{seg}_p$ is
	\[
	\operatorname{seg}_p=\left\{v \in T_{p} \mathbb{D}^n_r \mid \exp(t v):[0,1] \rightarrow  \mathbb{D}^n_r \text { is a segment }\right\},
	\]
	which satisfies $\mathbb{D}^n_r=\exp(\operatorname{seg}_p)$.
\end{definition}

For the Riemannian exponential map, Definition~\ref{thm2} indicates that the segment domain $\operatorname{seg}_p$ is a closed star-shaped subset of $\mathbb{R}^n$.
As for the exponential parametrization cluster, we have $\mathbb{D}^n_r=\phi_{\mathcal{F}_1}(\operatorname{seg}^*_p) \cup \phi_{\mathcal{F}_2}(\operatorname{seg}^*_p) \cdots \cup \phi_{\mathcal{F}_t}(\operatorname{seg}^*_p)$. This implies that, in order to cover $\mathbb{D}^n_r$, the required segment $\operatorname{seg}^*_p$ for the exponential parametrization cluster is less than or equal to the required segment $\operatorname{seg}_p$ for the Riemannian exponential map, i.e., $\operatorname{seg}^*_p \subseteq  \operatorname{seg}_p$. In practice, the exponential parametrization cluster increases the probability of weight flips by shrinking the segment domain, suggesting that weight vectors in $\operatorname{seg}^*_p$ can explore more efficiently than in $\operatorname{seg}_p$.

\subsection{Method Comparison and Explanation}

\textbf{HBNN} \emph{vs.} \textbf{BNN.} The improvement of HBNN over the general BNN can be primarily attributed to the unconstrained optimization via the exponential parametrization cluster.
In back-propagation, HBNN introduces additional computational overhead to the training process.
Considering Eq.(\ref{p}) and Eq.(\ref{w}), we update both $\mathcal{F}$ and $\tilde{\mathbf{w}}$, thereby increasing the number of trainable parameters.
In the inference phase, HBNN behaves similarly to general BNNs because both $\tilde{\mathbf{w}}$ and $\mathcal{F}$ contribute to binarized weight vectors $\operatorname{sign}(\mathbf{w})=\operatorname{sign}(\phi_{\mathcal{F}_i}(\tilde{\mathbf{w}}))$ based on the optimal exponential parametrization $\phi_{\mathcal{F}_i}$. While general BNNs obtain binarized weight vectors through $\operatorname{sign}(\tilde{\mathbf{w}})$, the representations of $\operatorname{sign}(\tilde{\mathbf{w}})$ and $\operatorname{sign}(\mathbf{w})$ involve the same parameter size and OPs in the inference phase. Therefore, HBNN does not introduce additional computational overhead to the inference process.

\textbf{HBNN} \emph{vs.} \textbf{MD.} The method of MD~\cite{ajanthan2021mirror} presents the mirror descent framework, mapping variables from the unconstrained space to the quantized one, which proves beneficial for BNN optimization. 
In contrast, HBNN provides the Riemannian geometry framework, mapping variables from the unconstrained space to hyperbolic space.
From the perspective of mapping, the mirror map of MD is set artificially, whereas the exponential parametrization cluster in HBNN is optimized via the derivative of the loss function with respect to $\mathcal{F}$.
From the perspective of optimization, the unconstrained problem of MD solely aims at optimizing the weight vector.
However, HBNN also takes into account the optimization of the mapping itself, i.e., the exponential parametrization cluster. This dual optimization is advantageous for increasing the probability of weight flips to maximize the information gain.

\section{Experiments}

\renewcommand{\baselinestretch}{1.2}
\begin{table}[thbp]
	\caption{Top-1 classification accuracy results on CIFAR100 with ResNet18 w.r.t. different radius $r$.}
	\begin{center}
		\begin{tabular}{ccccc}
			\cline{1-2} \cline{4-5}
			\multicolumn{2}{c}{\textbf{Parametr space} ($\mathbb{D}_{r}^{n}$)} & & \multicolumn{2}{c}{\textbf{Parametr space} ($\mathbb{S}_{r}^{n}$)} \\
			\cline{1-2} \cline{4-5}
			\text{Radius} & \text{mean} $\pm$ \text{std} (\%) & & \text{Radius} & \text{mean} $\pm$ \text{std} (\%) \\
			\cline{1-2} \cline{4-5}
			0.01 & 69.34 $\pm$ 0.15 && 0.01 & 69.24 $\pm$ 0.10 \\
			\cline{1-2} \cline{4-5}
			\textbf{0.05} & \textbf{69.50 $\pm$ 0.10} && 0.05 & 69.31 $\pm$ 0.37 \\
			\cline{1-2} \cline{4-5}
			0.10 & 69.45 $\pm$ 0.09 && 0.10 & 68.96 $\pm$ 0.27 \\
			\cline{1-2} \cline{4-5}
			0.50 & 69.33 $\pm$ 0.19 && 0.50 & 69.16 $\pm$ 0.09 \\
			\cline{1-2} \cline{4-5}
			1.00 & 69.19 $\pm$ 0.21 && \textbf{1.00} & \textbf{69.47 $\pm$ 0.11} \\
			\cline{1-2} \cline{4-5}
			5.00 & 68.84 $\pm$ 0.33 && 5.00 & 69.01 $\pm$ 0.17 \\
			\cline{1-2} \cline{4-5}
		\end{tabular}
	\end{center}
	\label{table1}
\end{table}
\renewcommand{\baselinestretch}{1}

\begin{figure}[thbp]
	\centering
	\includegraphics[width=.49\textwidth]{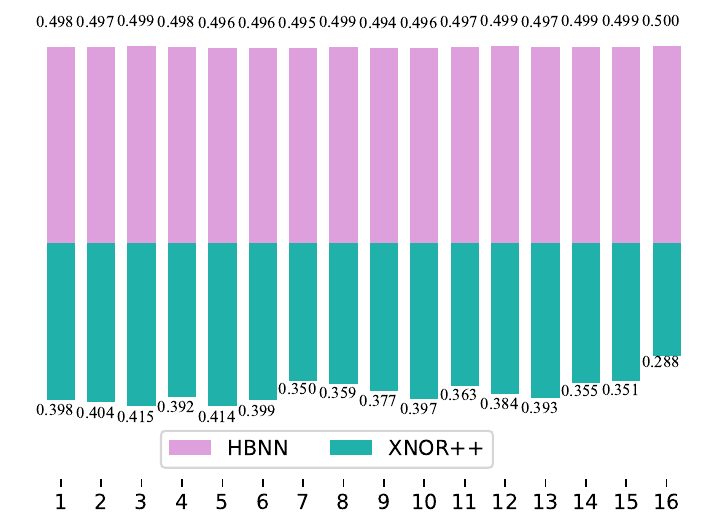}
	\caption{Weight flip rates of our HBNN and XNOR++ in different layers of ResNet18.}
	\label{fig3}
\end{figure}

\begin{figure}[thbp]
	\centering
	\includegraphics[width=.49\textwidth]{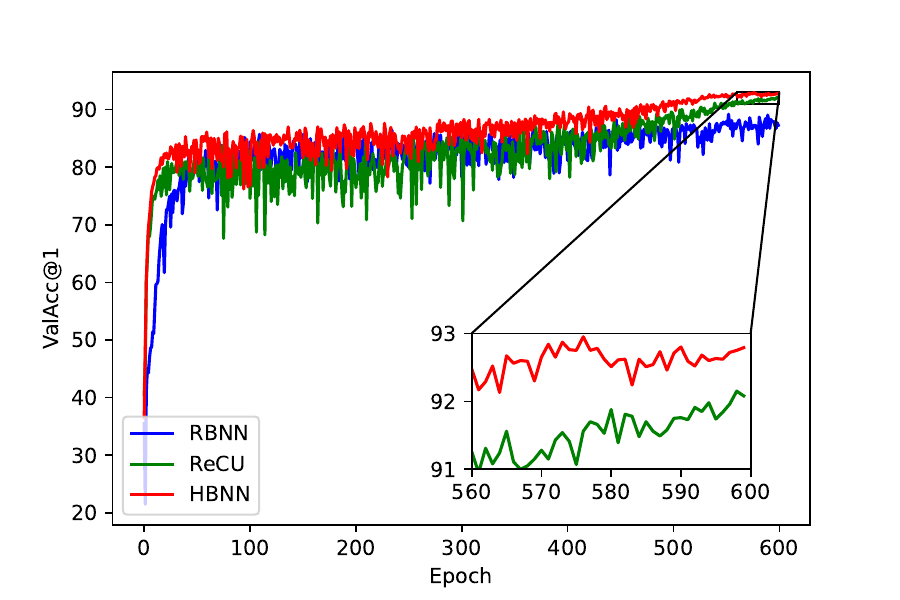}
	\caption{Validation accuracy curves of our HBNN, RBNN, and ReCU on CIFAR10 dataset with VGGsmall.}
	\label{curve}
\end{figure}

In this section, we conduct experiments to compare our HBNN, trained from scratch, with existing state-of-the-art methods in classification tasks. We evaluate the performance of the proposed method on CIFAR~\cite{krizhevsky2009learning} and ImageNet~\cite{krizhevsky2012imagenet} datasets.
All experiments are implemented on NVIDIA 3090Ti using the PyTorch framework.

\textbf{CIFAR datasets.} The CIFAR benchmarks consist of natural color images with 32x32 pixels. There are two datasets: CIFAR10 (C10) with images organized into 10 classes, and CIFAR100 (C100) with images organized into 100 classes. Each dataset comprises 50k training images and 10k test images. We adopt a standard data augmentation scheme, including random clipping and flipping, which is widely used~\cite{wang2021accelerate}. The images are normalized in preprocessing using the means and standard deviations of channels.

\begin{table}[t]
	\caption{Top-1 classification accuracy results on CIFAR10 and CIFAR100 datasets with ResNet18 and VGGsmall. W/A denotes the bit-width of weights/activations.}
	\begin{center}
		\begin{tabular}{ccccc}
			\toprule
			& Method & \makecell[c]{Bit \\ (W/A)} & \makecell[c]{Acc.(\%) \\ (C10)} & \makecell[c]{Acc.(\%) \\ (C100)} \\
			\midrule
			\multirow{10}{*}{\rotatebox{90}{ResNet18}} & Full-precision & 32/32 & 94.8 & 77.0 \\
			& IR-Net~\cite{qin2020forward} & 1/1 & 91.5 & 64.5 \\
			& RBNN~\cite{lin2020rotated} & 1/1 & 92.2 & 65.3 \\
			& IR-Net+CMIM~\cite{shang2022network} & 1/1 & 92.2 & 71.2 \\
                & ReSTE~\cite{wu2023estimator} & 1/1 & 92.6 & - \\
			& ReCU~\cite{xu2021recu} & 1/1 & 92.8 & - \\
			& SBNN (ours) & 1/1 & 92.8 & 71.2 \\
			& HBNN (ours) & 1/1 & \textbf{93.3} & \textbf{71.7} \\ \cline{2-5}
			& BC~\cite{courbariaux2015binaryconnect} & 1/32 & 91.6 & 72.1 \\
			& MD-softmax-s~\cite{ajanthan2021mirror} & 1/32 & 93.3 & 72.2 \\
                & SBNN (ours) & 1/32 & unstable & unstable \\
			& HBNN (ours) & 1/32 & \textbf{94.8} & \textbf{74.8} \\ \midrule
			\multirow{12}{*}{\rotatebox{90}{VGGsmall}} & Full-precision & 32/32 & 94.1 & 75.5 \\
			& XNOR~\cite{rastegari2016xnor} & 1/1 & 89.8 & - \\
			& DoReFa~\cite{zhou2016dorefa} & 1/1 & 90.2 & - \\
			& RAD~\cite{ding2019regularizing} & 1/1 & 90.5 & - \\
			& Proxy-BNN~\cite{he2020proxybnn} & 1/1 & 91.8 & 67.2 \\
			& RBNN~\cite{lin2020rotated} & 1/1 & 91.3 & 67.4 \\
			& DSQ~\cite{gong2019differentiable} & 1/1 & 91.7 & - \\
			& SLB~\cite{yang2020searching} & 1/1 & 92.0 & - \\
			& ReCU~\cite{xu2021recu} & 1/1 & 92.2 & - \\
			& RBNN+CMIM~\cite{shang2022network} & 1/1 & 92.2 & 71.0 \\
                & ReSTE~\cite{wu2023estimator} & 1/1 & 92.5 & - \\
			& SBNN (ours) & 1/1 & 92.8 & 72.2 \\
			& HBNN (ours) & 1/1 & \textbf{93.4} & \textbf{72.6} \\
			\bottomrule
		\end{tabular}
	\end{center}
	\label{table2}
\end{table}

\begin{table}[t]
	\caption{Top-1 and Top-5 classification accuracy results on ImageNet dataset with ResNet18 and ResNet34. W/A denotes the bit-width of weights/activations.}
	\begin{center}
		\begin{tabular}{ccccc}
			\toprule
			& Method & \makecell[c]{Bit \\ (W/A)} & \makecell[c]{Acc.(\%) \\ (Top-1)} & \makecell[c]{Acc.(\%) \\ (Top-5)} \\
			\midrule
			\multirow{11}{*}{\rotatebox{90}{ResNet18}} & Full-precision & 32/32 & 69.6 & 89.2 \\
			& ABC-Net~\cite{lin2017towards} & 1/1 & 42.7 & 67.6 \\
			& XNOR~\cite{rastegari2016xnor} & 1/1 & 51.2 & 73.2 \\
			& BiReal~\cite{liu2020bi} & 1/1 & 56.4 & 79.5 \\
			& IR-Net~\cite{qin2020forward} & 1/1 & 58.1 & 80.0 \\
			& RBNN~\cite{lin2020rotated} & 1/1 & 59.9 & 81.9 \\
			& FDA-BNN~\cite{xu2021learning} & 1/1 & 60.2 & 82.3 \\
			& ReCU~\cite{xu2021recu} & 1/1 & 61.0 & 82.6 \\
			& RBNN+CMIM~\cite{shang2022network} & 1/1 & 61.2 & 82.2 \\
                & SBNN (ours) & 1/1 & 61.5 & 83.3 \\
                & ReBNN~\cite{xu2023resilient} & 1/1 & 61.6 & 83.4 \\
			& HBNN (ours) & 1/1 & \textbf{61.8} & \textbf{83.6} \\ \midrule
			\multirow{10}{*}{\rotatebox{90}{ResNet34}} & Full-precision & 32/32 & 73.3 & 91.3 \\
			& XNOR++~\cite{bulat2019xnor} & 1/1 & 57.1 & 79.9 \\
			& LNS~\cite{han2020training} & 1/1 & 59.4 & 81.7 \\
			& BiReal~\cite{liu2020bi} & 1/1 & 62.2 & 83.9 \\
			& IR-Net~\cite{qin2020forward} & 1/1 & 62.9 & 84.1 \\
			& RBNN~\cite{lin2020rotated} & 1/1 & 63.1 & 84.4 \\
			& RBNN+CMIM~\cite{shang2022network} & 1/1 & 65.0 & 85.7 \\
			& ReCU~\cite{xu2021recu} & 1/1 & 65.1 & 85.8 \\
			& SBNN (ours) & 1/1 & 65.6 & 86.0 \\
                & ReBNN~\cite{xu2023resilient} & 1/1 & 65.8 & 86.2 \\
			& HBNN (ours) & 1/1 & \textbf{65.9} & \textbf{86.4} \\
			\bottomrule
		\end{tabular}
	\end{center}
	\label{table3}
\end{table}

\textbf{ImageNet dataset.} The ImageNet benchmark consists of 1.2 million high-resolution natural images, with a validation set containing 50k images. These images are organized into 1000 object categories for training and resized to 224x224 pixels before being fed into the network. Standard data augmentation strategies, such as random clips and horizontal flips~\cite{wang2021accelerate}, are applied. Single-crop evaluation results are reported using Top-1 and Top-5 accuracies.

\textbf{Experimental Setup.} For CIFAR datasets, our HBNNs are trained for a total of 600 epochs with a batch size of 256. We adopt the SGD optimizer with a momentum of 0.9 and a weight decay of 5e-4.
For the ImageNet dataset, our HBNN is trained for a total of 250 epochs with a batch size of 512. The same SGD optimizer settting are used, with a momentum of 0.9 and a weight decay of 1e-4. Notably, we initialize the learning rate at 0.1 and utilize the cosine learning rate scheduler in CIFAR10/CIFAR100 and ImageNet.

\subsection{Ablation Study}

\begin{figure*}[t]
	\centering
	\begin{minipage}{0.32\linewidth}[XNOR++]
		\centering
		\includegraphics[width=1\linewidth]{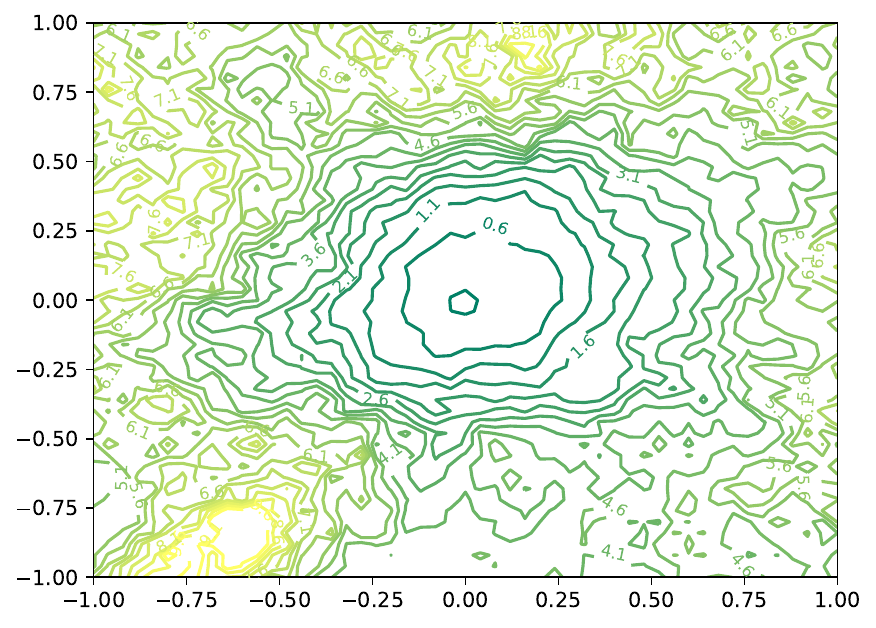}
		\label{fig:short-a}
	\end{minipage}
	\centering
	\begin{minipage}{0.32\linewidth}[HBNN]
		\centering
		\includegraphics[width=1\linewidth]{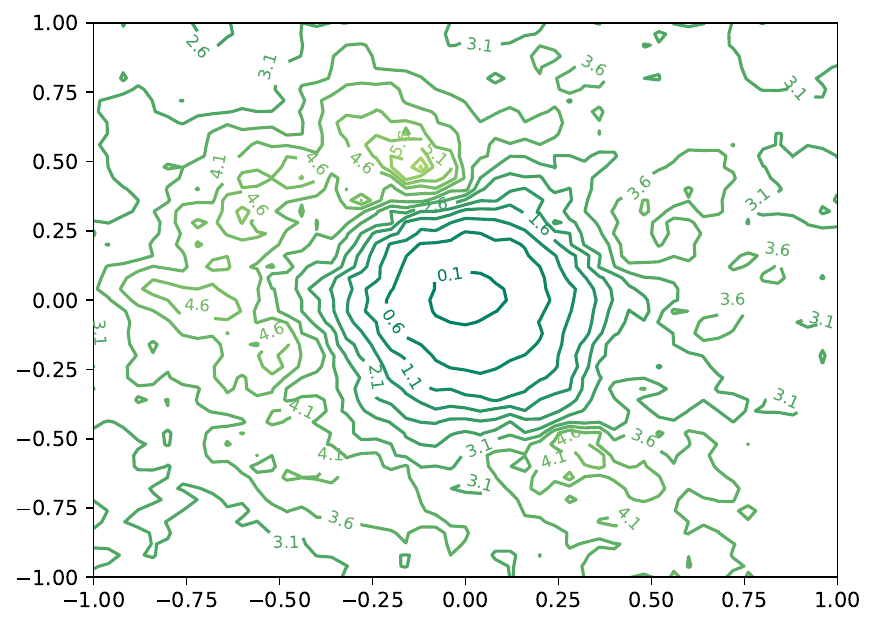}
		\label{fig:short-b}
	\end{minipage}
	\centering
	\begin{minipage}{0.32\linewidth}[Full-precision]
		\centering
		\includegraphics[width=1\linewidth]{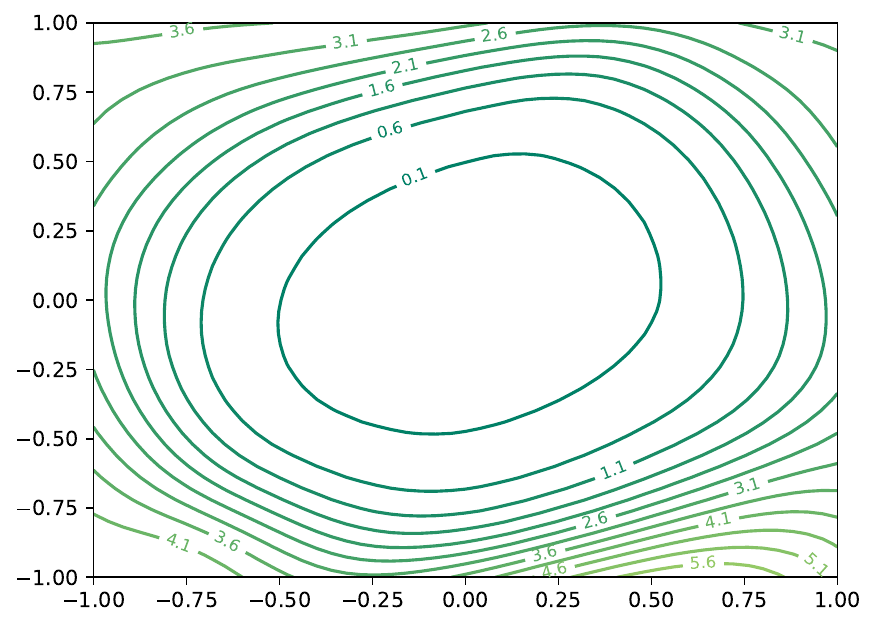}
		\label{fig:short-c}
	\end{minipage}
	\caption{2D visualization of the loss surfaces of ResNet18 on CIFAR10 dataset enables comparisons of the sharpness/flatness of different methods. The sharpness of loss surfaces is indicated by the accompanying numbers, with the yellow area representing particularly large peaks. In comparison to XNOR++, HBNN exhibits flatter loss surfaces.}
	\label{lossplane}
\end{figure*}

\begin{figure}[thbp]
	\centering
	\includegraphics[width=.49\textwidth]{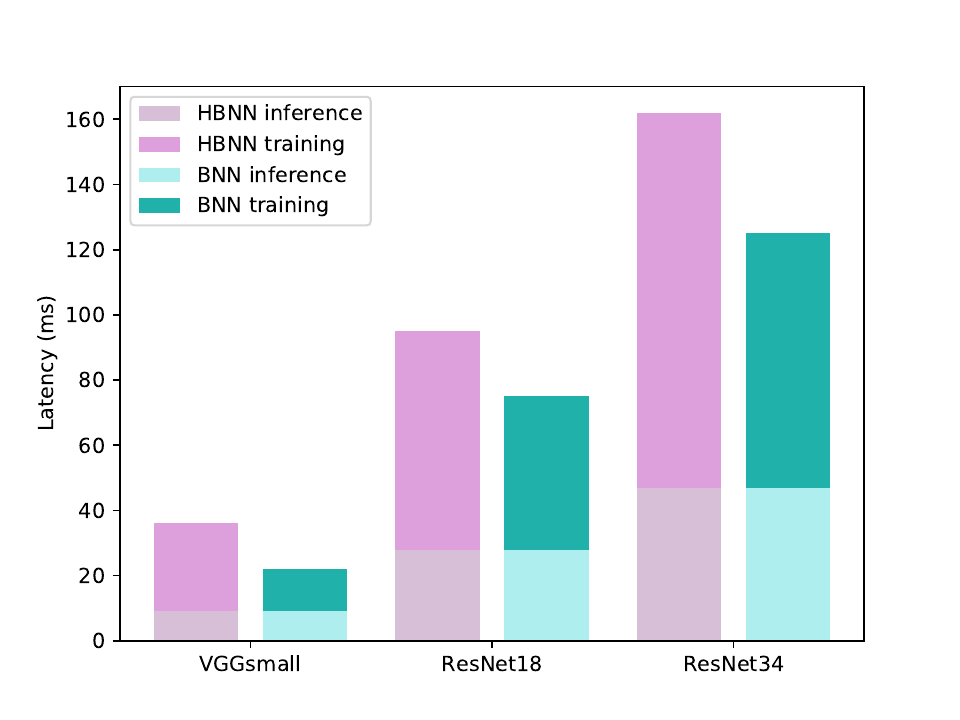}
	\caption{Latency comparison between HBNN and BNN during inference and training phases.}
	\label{latency}
\end{figure}

We conduct a series of ablation studies on CIFAR100 using the ResNet18 model. Leveraging two parameter spaces, namely HBNN for the Poincar\'e ball $\mathbb{D}_{r}^{n}$ and SBNN for the sphere $\mathbb{S}_{r}^{n}$, i.e., the boundary of the Poincar\'e ball $\partial\mathbb{D}_{r}^{n}$), we adjust different radius $r$ to determine the optimal radius by evaluating classification accuracies at epoch 120. The mean top-1 accuracies (mean $\pm$ std) are presented in Table~\ref{table1}. Consequently, we determine $r=0.05$ for the parameter space $\mathbb{D}_{r}^{n}$ and $r=1$ for the parameter space $\mathbb{S}_{r}^{n}$, which will be used in subsequent experiments. While the radius choice has a slight impact, its effect is minimal when considering the variation in results due to different random seeds. Thus, the radius can be considered a robust parameter in our method.

\begin{table}[htbp]
	\caption{The parameter size and OPs in ResNet models.}
	\begin{center}
		\begin{tabular}{ccccc}
			\toprule
			& Method & Size (MB)  & Size reduction & OPs ($10^8$) \\
			\midrule
			\multirow{5}{*}{\rotatebox{90}{ResNet18}} & Full-precision & 46.76 & - & 18.21  \\
			& ReBNN~\cite{xu2023resilient} & 4.15 & 11.26$\times$ & 1.63  \\
			& RBNN~\cite{lin2020rotated} & 4.15 & 11.26$\times$ & 1.63 \\
			& ReCU~\cite{xu2021recu} & 4.15 & 11.26$\times$ & 1.63 \\
			& HBNN (ours) & 4.15 & 11.26$\times$ & 1.63 \\ \midrule
			\multirow{5}{*}{\rotatebox{90}{ResNet34}} & Full-precision & 87.19 & - & 36.74  \\
			& ReBNN~\cite{xu2023resilient} & 5.41 & 16.11$\times$ & 1.93  \\
			& RBNN~\cite{lin2020rotated} & 5.41 & 16.11$\times$ & 1.93 \\
			& ReCU~\cite{xu2021recu} & 5.41 & 16.11$\times$ & 1.93 \\
			& HBNN (ours) & 5.41 & 16.11$\times$ & 1.93 \\
			\bottomrule
		\end{tabular}
	\end{center}
	\label{table6}
\end{table}

Figure~\ref{fig3} illustrates the weight flip rates of our HBNN and XNOR++ in different layers of ResNet18 on CIFAR10. As observed, HBNN results in approximately 50\% weight flips in each layer, demonstrating that the exponential parametrization cluster effectively increases the probability of weight flips.

\subsection{Comparison to State-of-the-art Methods}

The validation curves for ResNet18 are presented in Figure~\ref{curve}. In comparison to RBNN~\cite{lin2020rotated} and ReCU~\cite{xu2021recu} on CIFAR10, the validation accuracies of our HBNN show robust and stable convergence throughout the training epochs.

We conduct a thorough evaluation of our method against state-of-the-art methods, repeating each experiment 5 times and reporting statistics from the last 10/5 epochs' test accuracies for a fair comparison. As indicated in Table~\ref{table2}, HBNN consistently outperforms existing SOTA methods. Notably, our HBNN (with 1-bit weights and 1-bit activations) achieves performance improvements exceeding 1.2\% and 1.6\% with the VGGsmall architecture on CIFAR10 and CIFAR100, respectively. 
For the 1/32 case, the instability in the training of SBNN is noteworthy, possibly stemming from the exponential parametrization cluster stopping a diffeomorphism in the sphere $\mathbb{S}^n_r$, based on our theoretical analysis in Section~\ref{sec5}.

Table~\ref{table3} highlights that HBNN consistently outperforms existing state-of-the-art methods in both top-1 and top-5 accuracies. Specifically, our proposed method achieves a 0.8\% improvement in top-1 accuracy with the ResNet18 and ResNet34 architectures compared to ReCU method on ImageNet.

\begin{table}[h]
	\caption{Top-1 classification accuracy results on CIFAR10 dataset with ResNet18 and VGGsmall. W/A denotes the bit-width of weights/activations.}
	\begin{center}
		\begin{tabular}{cccc}
			\toprule
			& Method & \makecell[c]{Bit-width \\ (W/A)} & \makecell[c]{Acc.(\%) \\ (CIFAR10)} \\
			\midrule
			\multirow{5}{*}{\rotatebox{90}{ResNet18}} & Full-precision & 32/32 & 94.8  \\
			& IR-Net~\cite{qin2020forward} & 1/1 & 91.5  \\
			& IR-Net+HBNN & 1/1 & 92.1 \\
			& ReCU~\cite{xu2021recu} & 1/1 & 92.8 \\
			& ReCU+HBNN  & 1/1 & 93.0 \\ \midrule
			\multirow{5}{*}{\rotatebox{90}{VGGsmall}} & Full-precision & 32/32 & 94.1 \\
			& IR-Net~\cite{qin2020forward} & 1/1 & 90.4  \\
			& IR-Net+HBNN & 1/1 & 92.6 \\
			& ReCU~\cite{xu2021recu} & 1/1 & 92.2 \\
			& ReCU+HBNN  & 1/1 & 93.0 \\
			\bottomrule
		\end{tabular}
	\end{center}
	\label{table4}
\end{table}

\begin{table}[h]
	\caption{Top-1 and Top-5 classification accuracy results on ImageNet dataset with ResNet18 and ResNet34. W/A denotes the bit-width of weights/activations.}
	\begin{center}
		\begin{tabular}{ccccc}
			\toprule
			& Method & \makecell[c]{Bit-width \\ (W/A)} & \makecell[c]{Acc.(\%) \\ (Top-1)} & \makecell[c]{Acc.(\%) \\ (Top-5)} \\
			\midrule
			\multirow{5}{*}{\rotatebox{90}{ResNet18}} & Full-precision & 32/32 & 69.6 & 89.2 \\
			& IR-Net~\cite{qin2020forward} & 1/1 & 58.1 & 80.0  \\
			& IR-Net+HBNN & 1/1 & 60.9 & 82.9 \\
			& ReCU~\cite{xu2021recu} & 1/1 & 61.0 & 82.6 \\
			& ReCU+HBNN  & 1/1 & 61.5 & 83.3 \\ \midrule
			\multirow{5}{*}{\rotatebox{90}{ResNet34}} & Full-precision & 32/32 & 73.3 & 91.3 \\
			& IR-Net~\cite{qin2020forward} & 1/1 & 62.9 & 84.1  \\
			& IR-Net+HBNN & 1/1 & 64.2 & 85.2 \\
			& ReCU~\cite{xu2021recu} & 1/1 & 65.1 & 85.8 \\
			& ReCU+HBNN  & 1/1 & 65.8 & 86.4 \\
			\bottomrule
		\end{tabular}
	\end{center}
	\label{table5}
\end{table}

\subsection{Visualization}
\label{app3}

Additionally, we provide a 2D visualization of the loss surfaces for both HBNN and XNOR++ in according with previous work~\cite{li2018visualizing}. Analyzing Figure~\ref{lossplane}, it becomes evident that the loss surface of the full-precision model is smooth and flat, a characteristic beneficial for training and representing in neural networks. With the exponential parametrization cluster, HBNN maintains the property of diffeomorphism by not introducing or eliminating local minima in the loss surfaces, in contrast to XNOR++. Consequently, HBNN exhibits relatively flatter loss surfaces than XNOR++, suggesting that the exponential parametrization cluster contributes to more effective optimization of BNNs.

\subsection{Computational Complexity}
Based on the analysis in Section~\ref{sec5}, HBNN introduces additional computational overhead to the training process as it requires training weights and the exponential parametrization cluster. Nevertheless, during inference, HBNN behaves similarly to general BNNs because both $\tilde{\mathbf{w}}$ and $\mathcal{F}$ contribute to binarized weight vectors $\operatorname{sign}(\mathbf{w})=\operatorname{sign}(\phi_{\mathcal{F}_i}(\tilde{\mathbf{w}}))$ based on the optimal exponential parametrization $\phi_{\mathcal{F}_i}$. While general BNNs obtain binarized weight vectors through $\operatorname{sign}(\tilde{\mathbf{w}})$, the representations of $\operatorname{sign}(\tilde{\mathbf{w}})$ and $\operatorname{sign}(\mathbf{w})$ involve the same parameter size and OPs in the inference phase. Therefore, HBNN does not introduce additional computational overhead to the inference process.

In Figure~\ref{latency}, when comparing the latency of HBNN and BNN in the inference and training phases, we observe that the latency of HBNN is slightly higher than that of BNN during the training process across different models, while their inference latency remain consistent, thereby confirming our previous analysis. Furthermore, we use the parameter size and OPs following~\cite{xu2023resilient} for comparison with other methods. As shown in Table~\ref{table6}, HBNN exhibits the same parameter size and OPs as other methods in the inference phase, which is significant for real-time applications.

\subsection{Compatibility}
\label{app1}

We further evaluate the compatibility of HBNN to illustrate the universality of our method.
We integrate HBNN into IR-Net and ReCU as a plug-and-play module, as presented in Table~\ref{table4} and~\ref{table5}.
The incorporation of HBNN into these methods results in a noticeable performance improvement.

\section{Conclusion}

This paper introduces the optimization framework of HBNN, which transforms a constrained problem in hyperbolic space into an unconstrained one in Euclidean space using the exponential parametrization cluster.
Through the analysis of the exponential parametrization cluster, we have determined that it accelerates the exploration of weight vectors, thereby increasing the probability of weight flips compared to the Riemannian exponential map.
Experimental results demonstrate that HBNN achieves approximately 50\% weight flips, effectively optimizing BNNs to achieve state-of-the-art performance.
In the future, our focus will shift towards further exploring the optimization of neural networks from a geometrical perspective.


\bibliographystyle{Bibliography/IEEEtranTIE}
\bibliography{Bibliography/IEEEabrv,Bibliography/BIB_1x-TIE-2xxx}\ 

\begin{thebibliography}{10}
\providecommand{\url}[1]{#1}
\csname url@samestyle\endcsname
\providecommand{\newblock}{\relax}
\providecommand{\bibinfo}[2]{#2}
\providecommand{\BIBentrySTDinterwordspacing}{\spaceskip=0pt\relax}
\providecommand{\BIBentryALTinterwordstretchfactor}{4}
\providecommand{\BIBentryALTinterwordspacing}{\spaceskip=\fontdimen2\font plus
\BIBentryALTinterwordstretchfactor\fontdimen3\font minus
  \fontdimen4\font\relax}
\providecommand{\BIBforeignlanguage}[2]{{%
\expandafter\ifx\csname l@#1\endcsname\relax
\typeout{** WARNING: IEEEtran.bst: No hyphenation pattern has been}%
\typeout{** loaded for the language `#1'. Using the pattern for}%
\typeout{** the default language instead.}%
\else
\language=\csname l@#1\endcsname
\fi
#2}}
\providecommand{\BIBdecl}{\relax}
\BIBdecl

\bibitem{krizhevsky2012imagenet}
A.~Krizhevsky, I.~Sutskever, and G.~E. Hinton, ``Imagenet classification with
  deep convolutional neural networks,'' \emph{Advances in neural information
  processing systems}, vol.~25, 2012.

\bibitem{he2016deep}
K.~He, X.~Zhang, S.~Ren, and J.~Sun, ``Deep residual learning for image
  recognition,'' in \emph{Proceedings of the IEEE conference on computer vision
  and pattern recognition}, pp. 770--778, 2016.

\bibitem{redmon2016you}
J.~Redmon, S.~Divvala, R.~Girshick, and A.~Farhadi, ``You only look once:
  Unified, real-time object detection,'' in \emph{Proceedings of the IEEE
  conference on computer vision and pattern recognition}, pp. 779--788, 2016.

\bibitem{he2017mask}
K.~He, G.~Gkioxari, P.~Doll{\'a}r, and R.~Girshick, ``Mask r-cnn,'' in
  \emph{Proceedings of the IEEE international conference on computer vision},
  pp. 2961--2969, 2017.

\bibitem{long2015fully}
J.~Long, E.~Shelhamer, and T.~Darrell, ``Fully convolutional networks for
  semantic segmentation,'' in \emph{Proceedings of the IEEE conference on
  computer vision and pattern recognition}, pp. 3431--3440, 2015.

\bibitem{noh2015learning}
H.~Noh, S.~Hong, and B.~Han, ``Learning deconvolution network for semantic
  segmentation,'' in \emph{Proceedings of the IEEE international conference on
  computer vision}, pp. 1520--1528, 2015.

\bibitem{ding2019global}
X.~Ding, X.~Zhou, Y.~Guo, J.~Han, J.~Liu \emph{et~al.}, ``Global sparse
  momentum sgd for pruning very deep neural networks,'' \emph{Advances in
  Neural Information Processing Systems}, vol.~32, 2019.

\bibitem{lin2020hrank}
M.~Lin, R.~Ji, Y.~Wang, Y.~Zhang, B.~Zhang, Y.~Tian, and L.~Shao, ``Hrank:
  Filter pruning using high-rank feature map,'' in \emph{Proceedings of the
  IEEE/CVF conference on computer vision and pattern recognition}, pp.
  1529--1538, 2020.

\bibitem{bai2023unified}
S.~Bai, J.~Chen, X.~Shen, Y.~Qian, and Y.~Liu, ``Unified data-free compression:
  Pruning and quantization without fine-tuning,'' in \emph{Proceedings of the
  IEEE/CVF International Conference on Computer Vision}, pp. 5876--5885, 2023.

\bibitem{banner2018scalable}
R.~Banner, I.~Hubara, E.~Hoffer, and D.~Soudry, ``Scalable methods for 8-bit
  training of neural networks,'' \emph{Advances in neural information
  processing systems}, vol.~31, 2018.

\bibitem{helwegen2019latent}
K.~Helwegen, J.~Widdicombe, L.~Geiger, Z.~Liu, K.-T. Cheng, and R.~Nusselder,
  ``Latent weights do not exist: Rethinking binarized neural network
  optimization,'' \emph{Advances in neural information processing systems},
  vol.~32, 2019.

\bibitem{chen2020propagating}
J.~Chen, Y.~Liu, H.~Zhang, S.~Hou, and J.~Yang, ``Propagating
  asymptotic-estimated gradients for low bitwidth quantized neural networks,''
  \emph{IEEE Journal of Selected Topics in Signal Processing}, vol.~14, no.~4,
  pp. 848--859, 2020.

\bibitem{chen2023learning}
J.~Chen, H.~Chen, M.~Wang, G.~Dai, I.~W. Tsang, and Y.~Liu, ``Learning
  discretized neural networks under ricci flow,'' \emph{arXiv preprint
  arXiv:2302.03390}, 2023.

\bibitem{chen2023data}
J.~Chen, S.~Bai, T.~Huang, M.~Wang, G.~Tian, and Y.~Liu, ``Data-free
  quantization via mixed-precision compensation without fine-tuning,''
  \emph{Pattern Recognition}, p. 109780, 2023.

\bibitem{liu2023dccd}
Y.~Liu, J.~Chen, and Y.~Liu, ``Dccd: Reducing neural network redundancy via
  distillation,'' \emph{IEEE Transactions on Neural Networks and Learning
  Systems}, 2023.

\bibitem{9056829}
J.~Chen, L.~Liu, Y.~Liu, and X.~Zeng, ``A learning framework for n-bit
  quantized neural networks toward fpgas,'' \emph{IEEE Transactions on Neural
  Networks and Learning Systems}, vol.~32, no.~3, pp. 1067--1081, 2021.

\bibitem{bubeck2015convex}
S.~Bubeck \emph{et~al.}, ``Convex optimization: Algorithms and complexity,''
  \emph{Foundations and Trends{\textregistered} in Machine Learning}, vol.~8,
  no. 3-4, pp. 231--357, 2015.

\bibitem{ajanthan2021mirror}
T.~Ajanthan, K.~Gupta, P.~Torr, R.~Hartley, and P.~Dokania, ``Mirror descent
  view for neural network quantization,'' in \emph{International Conference on
  Artificial Intelligence and Statistics}, pp. 2809--2817.\hskip 1em plus 0.5em
  minus 0.4em\relax PMLR, 2021.

\bibitem{lin2020rotated}
M.~Lin, R.~Ji, Z.~Xu, B.~Zhang, Y.~Wang, Y.~Wu, F.~Huang, and C.-W. Lin,
  ``Rotated binary neural network,'' \emph{Advances in neural information
  processing systems}, vol.~33, pp. 7474--7485, 2020.

\bibitem{guggenheimer2012differential}
H.~W. Guggenheimer, \emph{Differential geometry}.\hskip 1em plus 0.5em minus
  0.4em\relax Courier Corporation, 2012.

\bibitem{absil2009optimization}
P.-A. Absil, R.~Mahony, and R.~Sepulchre, ``Optimization algorithms on matrix
  manifolds,'' in \emph{Optimization Algorithms on Matrix Manifolds}.\hskip 1em
  plus 0.5em minus 0.4em\relax Princeton University Press, 2009.

\bibitem{chen2024decentralized}
\BIBentryALTinterwordspacing
J.~Chen, H.~Ye, M.~Wang, T.~Huang, G.~Dai, I.~Tsang, and Y.~Liu,
  ``Decentralized riemannian conjugate gradient method on the stiefel
  manifold,'' in \emph{The Twelfth International Conference on Learning
  Representations}, 2024. [Online]. Available:
  \url{https://openreview.net/forum?id=PQbFUMKLFp}
\BIBentrySTDinterwordspacing

\bibitem{helfrich2018orthogonal}
K.~Helfrich, D.~Willmott, and Q.~Ye, ``Orthogonal recurrent neural networks
  with scaled cayley transform,'' in \emph{International Conference on Machine
  Learning}, pp. 1969--1978.\hskip 1em plus 0.5em minus 0.4em\relax PMLR, 2018.

\bibitem{lezcano2019cheap}
M.~Lezcano-Casado and D.~Mart{\i}nez-Rubio, ``Cheap orthogonal constraints in
  neural networks: A simple parametrization of the orthogonal and unitary
  group,'' in \emph{International Conference on Machine Learning}, pp.
  3794--3803.\hskip 1em plus 0.5em minus 0.4em\relax PMLR, 2019.

\bibitem{lezcano2019trivializations}
M.~Lezcano~Casado, ``Trivializations for gradient-based optimization on
  manifolds,'' \emph{Advances in Neural Information Processing Systems},
  vol.~32, 2019.

\bibitem{rastegari2016xnor}
M.~Rastegari, V.~Ordonez, J.~Redmon, and A.~Farhadi, ``Xnor-net: Imagenet
  classification using binary convolutional neural networks,'' in
  \emph{European conference on computer vision}, pp. 525--542.\hskip 1em plus
  0.5em minus 0.4em\relax Springer, 2016.

\bibitem{bulat2019xnor}
A.~Bulat and G.~Tzimiropoulos, ``Xnor-net++: Improved binary neural networks,''
  \emph{arXiv preprint arXiv:1909.13863}, 2019.

\bibitem{liu2020bi}
Z.~Liu, W.~Luo, B.~Wu, X.~Yang, W.~Liu, and K.-T. Cheng, ``Bi-real net:
  Binarizing deep network towards real-network performance,''
  \emph{International Journal of Computer Vision}, vol. 128, no.~1, pp.
  202--219, 2020.

\bibitem{he2020proxybnn}
X.~He, Z.~Mo, K.~Cheng, W.~Xu, Q.~Hu, P.~Wang, Q.~Liu, and J.~Cheng,
  ``Proxybnn: Learning binarized neural networks via proxy matrices,'' in
  \emph{European Conference on Computer Vision}, pp. 223--241.\hskip 1em plus
  0.5em minus 0.4em\relax Springer, 2020.

\bibitem{qin2020forward}
H.~Qin, R.~Gong, X.~Liu, M.~Shen, Z.~Wei, F.~Yu, and J.~Song, ``Forward and
  backward information retention for accurate binary neural networks,'' in
  \emph{Proceedings of the IEEE/CVF conference on computer vision and pattern
  recognition}, pp. 2250--2259, 2020.

\bibitem{xu2021recu}
Z.~Xu, M.~Lin, J.~Liu, J.~Chen, L.~Shao, Y.~Gao, Y.~Tian, and R.~Ji, ``Recu:
  Reviving the dead weights in binary neural networks,'' in \emph{Proceedings
  of the IEEE/CVF international conference on computer vision}, pp. 5198--5208,
  2021.

\bibitem{salimans2016weight}
T.~Salimans and D.~P. Kingma, ``Weight normalization: A simple
  reparameterization to accelerate training of deep neural networks,''
  \emph{Advances in neural information processing systems}, vol.~29, 2016.

\bibitem{huang2017centered}
L.~Huang, X.~Liu, Y.~Liu, B.~Lang, and D.~Tao, ``Centered weight normalization
  in accelerating training of deep neural networks,'' in \emph{Proceedings of
  the IEEE International Conference on Computer Vision}, pp. 2803--2811, 2017.

\bibitem{petersen2006riemannian}
P.~Petersen, \emph{Riemannian geometry}, vol. 171.\hskip 1em plus 0.5em minus
  0.4em\relax Springer, 2006.

\bibitem{courbariaux2016binarized}
M.~Courbariaux, I.~Hubara, D.~Soudry, R.~El-Yaniv, and Y.~Bengio, ``Binarized
  neural networks: Training deep neural networks with weights and activations
  constrained to+ 1 or-1,'' \emph{arXiv preprint arXiv:1602.02830}, 2016.

\bibitem{hinton2012neural}
G.~Hinton, N.~Srivastava, and K.~Swersky, ``Neural networks for machine
  learning,'' \emph{Coursera, video lectures}, vol. 264, no.~1, pp. 2146--2153,
  2012.

\bibitem{bengio2013estimating}
Y.~Bengio, N.~L{\'e}onard, and A.~Courville, ``Estimating or propagating
  gradients through stochastic neurons for conditional computation,''
  \emph{arXiv preprint arXiv:1308.3432}, 2013.

\bibitem{anderson2006hyperbolic}
J.~W. Anderson, \emph{Hyperbolic geometry}.\hskip 1em plus 0.5em minus
  0.4em\relax Springer Science \& Business Media, 2006.

\bibitem{ganea2018hyperbolic}
O.~Ganea, G.~B{\'e}cigneul, and T.~Hofmann, ``Hyperbolic neural networks,''
  \emph{Advances in neural information processing systems}, vol.~31, 2018.

\bibitem{nickel2017poincare}
M.~Nickel and D.~Kiela, ``Poincar{\'e} embeddings for learning hierarchical
  representations,'' \emph{Advances in neural information processing systems},
  vol.~30, 2017.

\bibitem{ganea2018hyperbolic1}
O.~Ganea, G.~B{\'e}cigneul, and T.~Hofmann, ``Hyperbolic entailment cones for
  learning hierarchical embeddings,'' in \emph{International Conference on
  Machine Learning}, pp. 1646--1655.\hskip 1em plus 0.5em minus 0.4em\relax
  PMLR, 2018.

\bibitem{ungar2001hyperbolic}
A.~A. Ungar, ``Hyperbolic trigonometry and its application in the poincar{\'e}
  ball model of hyperbolic geometry,'' \emph{Computers \& Mathematics with
  Applications}, vol.~41, no. 1-2, pp. 135--147, 2001.

\bibitem{ungar2008gyrovector}
A.~A. Ungar, ``A gyrovector space approach to hyperbolic geometry,''
  \emph{Synthesis Lectures on Mathematics and Statistics}, vol.~1, no.~1, pp.
  1--194, 2008.

\bibitem{petersen2016riemannian}
P.~Petersen, ``Riemannian metrics,'' in \emph{Riemannian Geometry}, pp.
  1--39.\hskip 1em plus 0.5em minus 0.4em\relax Springer, 2016.

\bibitem{krizhevsky2009learning}
A.~Krizhevsky, G.~Hinton \emph{et~al.}, ``Learning multiple layers of features
  from tiny images,'' 2009.

\bibitem{wang2021accelerate}
W.~Wang, M.~Chen, S.~Zhao, L.~Chen, J.~Hu, H.~Liu, D.~Cai, X.~He, and W.~Liu,
  ``Accelerate cnns from three dimensions: a comprehensive pruning framework,''
  in \emph{International Conference on Machine Learning}, pp.
  10\,717--10\,726.\hskip 1em plus 0.5em minus 0.4em\relax PMLR, 2021.

\bibitem{shang2022network}
Y.~Shang, D.~Xu, Z.~Zong, and Y.~Yan, ``Network binarization via contrastive
  learning,'' \emph{arXiv preprint arXiv:2207.02970}, 2022.

\bibitem{wu2023estimator}
X.-M. Wu, D.~Zheng, Z.~Liu, and W.-S. Zheng, ``Estimator meets equilibrium
  perspective: A rectified straight through estimator for binary neural
  networks training,'' in \emph{Proceedings of the IEEE/CVF International
  Conference on Computer Vision}, pp. 17\,055--17\,064, 2023.

\bibitem{courbariaux2015binaryconnect}
M.~Courbariaux, Y.~Bengio, and J.-P. David, ``Binaryconnect: Training deep
  neural networks with binary weights during propagations,'' \emph{Advances in
  neural information processing systems}, vol.~28, 2015.

\bibitem{zhou2016dorefa}
S.~Zhou, Y.~Wu, Z.~Ni, X.~Zhou, H.~Wen, and Y.~Zou, ``Dorefa-net: Training low
  bitwidth convolutional neural networks with low bitwidth gradients,''
  \emph{arXiv preprint arXiv:1606.06160}, 2016.

\bibitem{ding2019regularizing}
R.~Ding, T.-W. Chin, Z.~Liu, and D.~Marculescu, ``Regularizing activation
  distribution for training binarized deep networks,'' in \emph{Proceedings of
  the IEEE/CVF Conference on Computer Vision and Pattern Recognition}, pp.
  11\,408--11\,417, 2019.

\bibitem{gong2019differentiable}
R.~Gong, X.~Liu, S.~Jiang, T.~Li, P.~Hu, J.~Lin, F.~Yu, and J.~Yan,
  ``Differentiable soft quantization: Bridging full-precision and low-bit
  neural networks,'' in \emph{Proceedings of the IEEE/CVF International
  Conference on Computer Vision}, pp. 4852--4861, 2019.

\bibitem{yang2020searching}
Z.~Yang, Y.~Wang, K.~Han, C.~Xu, C.~Xu, D.~Tao, and C.~Xu, ``Searching for
  low-bit weights in quantized neural networks,'' \emph{Advances in neural
  information processing systems}, vol.~33, pp. 4091--4102, 2020.

\bibitem{lin2017towards}
X.~Lin, C.~Zhao, and W.~Pan, ``Towards accurate binary convolutional neural
  network,'' \emph{Advances in neural information processing systems}, vol.~30,
  2017.

\bibitem{xu2021learning}
Y.~Xu, K.~Han, C.~Xu, Y.~Tang, C.~Xu, and Y.~Wang, ``Learning frequency domain
  approximation for binary neural networks,'' \emph{Advances in Neural
  Information Processing Systems}, vol.~34, pp. 25\,553--25\,565, 2021.

\bibitem{xu2023resilient}
S.~Xu, Y.~Li, T.~Ma, M.~Lin, H.~Dong, B.~Zhang, P.~Gao, and J.~Lu, ``Resilient
  binary neural network,'' in \emph{Proceedings of the AAAI Conference on
  Artificial Intelligence}, vol.~37, no.~9, pp. 10\,620--10\,628, 2023.

\bibitem{han2020training}
K.~Han, Y.~Wang, Y.~Xu, C.~Xu, E.~Wu, and C.~Xu, ``Training binary neural
  networks through learning with noisy supervision,'' in \emph{International
  Conference on Machine Learning}, pp. 4017--4026.\hskip 1em plus 0.5em minus
  0.4em\relax PMLR, 2020.

\bibitem{li2018visualizing}
H.~Li, Z.~Xu, G.~Taylor, C.~Studer, and T.~Goldstein, ``Visualizing the loss
  landscape of neural nets,'' \emph{Advances in neural information processing
  systems}, vol.~31, 2018.

\end{thebibliography}

\begin{IEEEbiography}[{\includegraphics[width=1in,height=1.25in,clip,keepaspectratio]{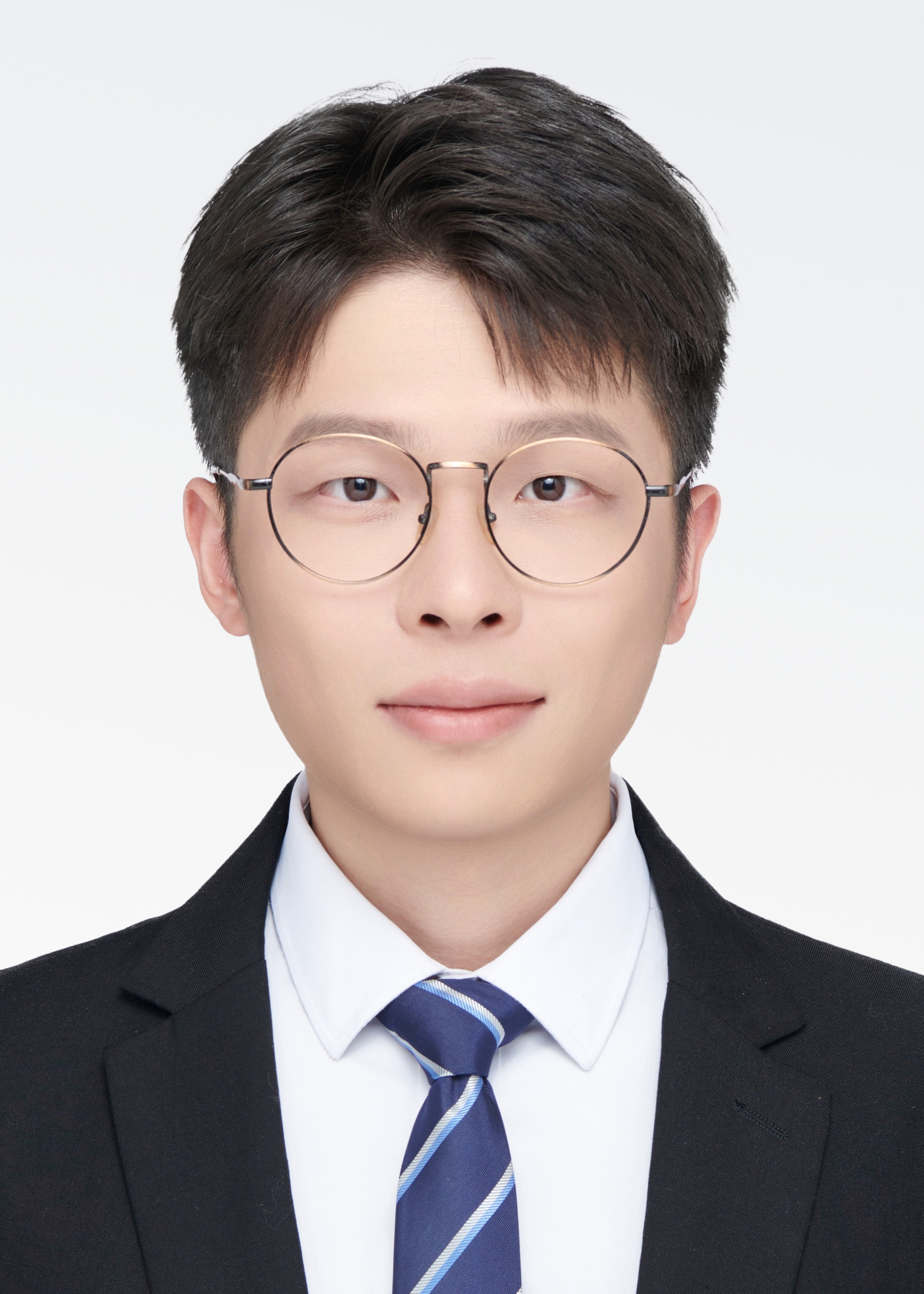}}]{Jun Chen} received the B.S. degree in the department of Mechanical and Electrical Engineering from China Jiliang University, Hangzhou, China, in 2016, and the M.S. degree in control engineering from the Zhejiang University, Hangzhou, China, in 2020, and the Ph.D degree in control science and engineering from Zhejiang University, Zhejiang, China, in 2024. He is currently a distinguished professor in Zhejiang Normal University. His research interests include deep learning, model compression, decentralized optimization, and manifold optimization.
\end{IEEEbiography}

\begin{IEEEbiography}[{\includegraphics[width=1in,height=1.25in,clip,keepaspectratio]{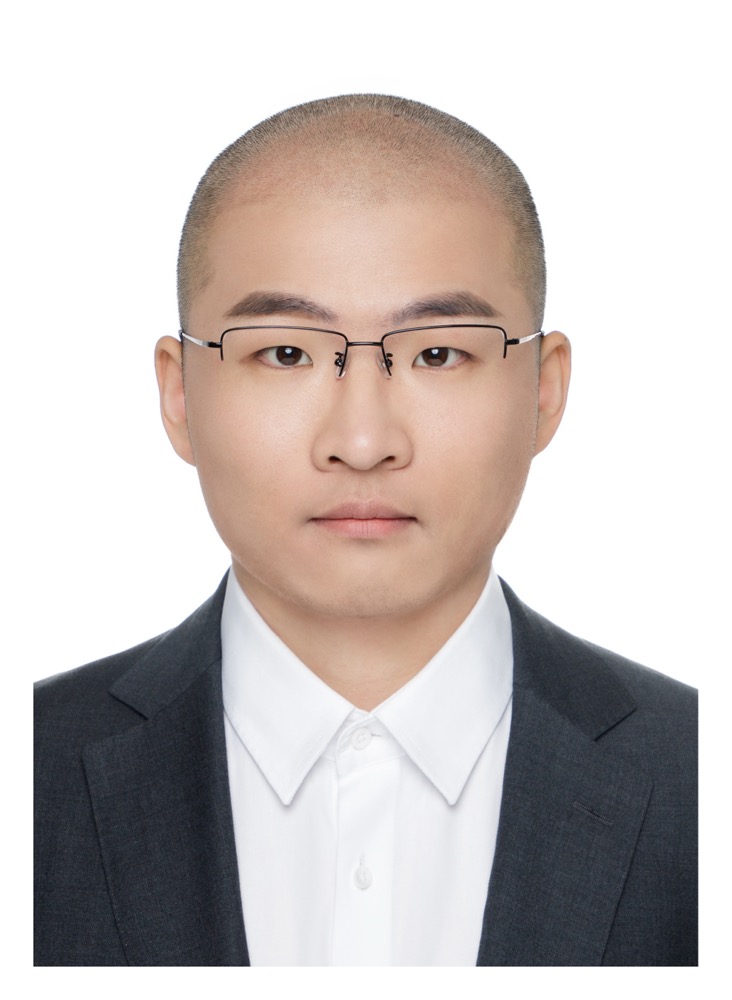}}]{Jingyang Xiang} received the B.S. degree in electrical engineering and automation from the Zhejiang University of Technology, Hangzhou, China, in 2022. He is pursuing his M.S. degree in College of Control Science and Engineering, Zhejiang University, Hangzhou, China. His current research interest is efficient AI, especially LLM quantization.
\end{IEEEbiography}

\begin{IEEEbiography}[{\includegraphics[width=1in,height=1.25in,clip,keepaspectratio]{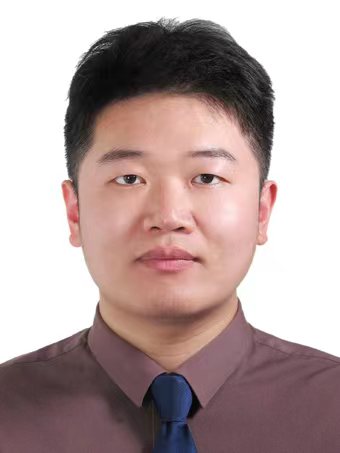}}]{Tianxin Huang} received the Bachelor degree in Mechanical Engineering from Xi'an Jiaotong University (XJTU) in 2017, and the Doctor's Degree at April Lab, Zhejiang University. He is currently a Research Fellow (Postdoc) in National University of Singapore (NUS), School of Computing, focusing on 3D Computer Vision. His current research interests includes but not limited on: 3D Reconstruction, Neural Rendering, 3D Face Reconstruction, 3D Point Cloud Analysis.
\end{IEEEbiography}

\begin{IEEEbiography}[{\includegraphics[width=1in,height=1.25in,clip,keepaspectratio]{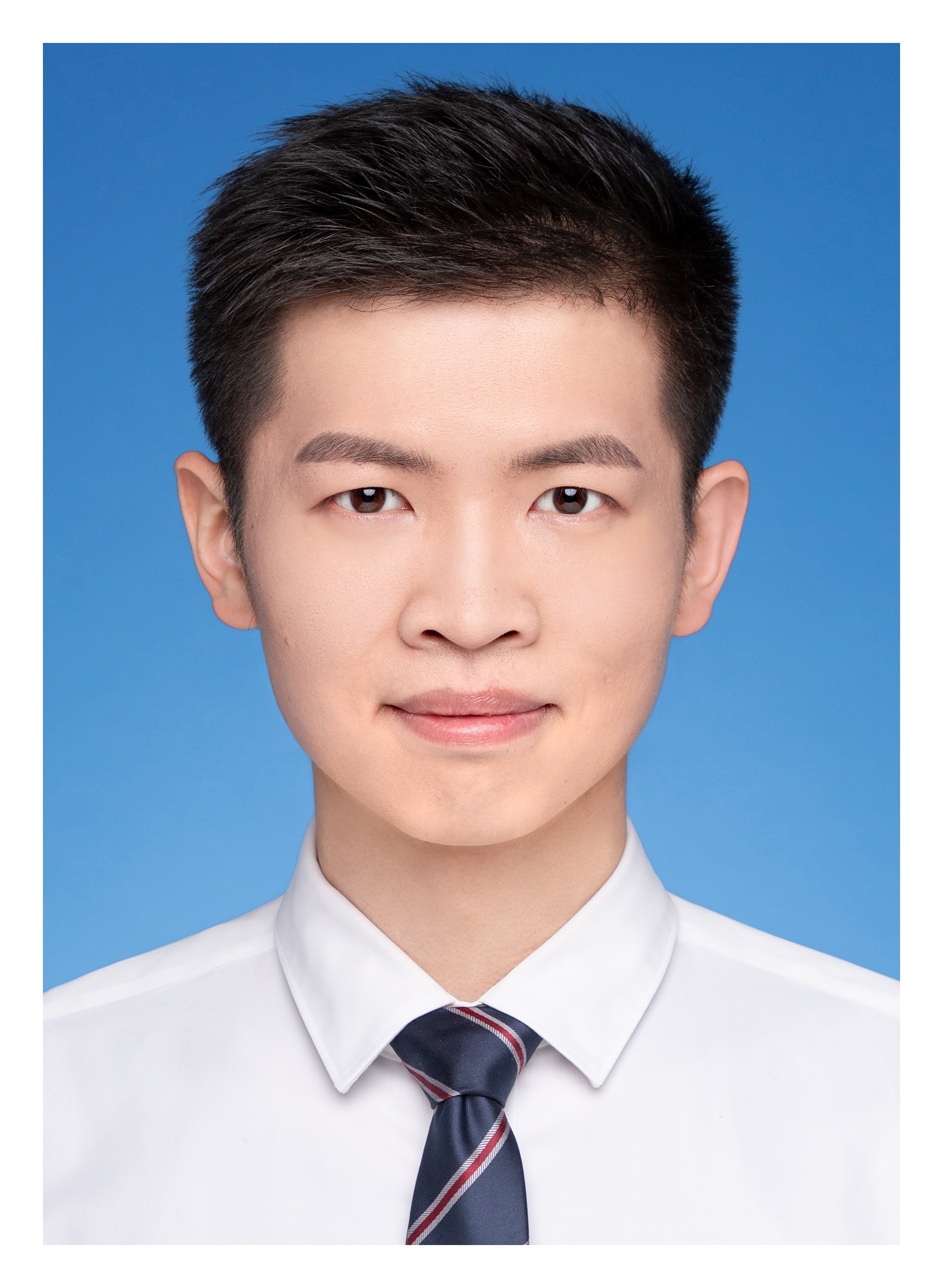}}]{Xiangrui Zhao} received the B.S. degree from Huazhong University of Science and Technology, Wuhan, China, in 2018, and his Ph.D. degree from the Institute of Cyber Systems and Control, Zhejiang University, Hangzhou, China, in 2023. His current research interests include BEV perception and end-to-end autonomous driving.
\end{IEEEbiography}

\begin{IEEEbiography}[{\includegraphics[width=1in,height=1.25in,clip,keepaspectratio]{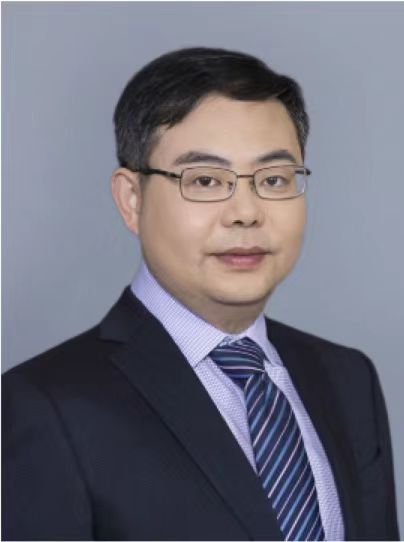}}]{Yong Liu} (Member, IEEE) received his B.S. degree in computer science and engineering from Zhejiang University in 2001, and the Ph.D. degree in computer science from Zhejiang University in 2007. He is currently a professor in the Institute of Cyber Systems and Control, Department of Control Science and Engineering, Zhejiang University. He has published more than 30 research papers in machine learning, computer vision, information fusion, robotics. His latest research interests include machine learning, robotics vision, information processing and granular computing.
\end{IEEEbiography}

\end{document}